\documentclass[letterpaper, 10 pt, journal, twoside]{IEEEtran} 

\usepackage{graphicx}
\usepackage{bm}
\usepackage{cite}
\usepackage[caption=false,font=footnotesize]{subfig}
\usepackage{multirow}
\usepackage{flushend}

\begin{document}

\title{Data-driven super resolution on a tactile dome}

\author{Pedro Piacenza$^*$, Sydney Sherman$^*$, and Matei Ciocarlie%
\thanks{Manuscript received: September, 10, 2017; Revised December, 8, 2017; Accepted January, 6, 2018.}
\thanks{This paper was recommended for publication by Editor John Wen upon evaluation of the Associate Editor and Reviewers' comments. 
This work was supported in part by NSF CAREER Award 1551631} 
\thanks{Department of Mechanical Engineering, Columbia University, \hfill\newline
New York, NY 10027, USA.\hfill\newline
{\tt\small pp2511@columbia.edu, ss4771@columbia.edu, matei.ciocarlie@columbia.edu}}%
\thanks{$^*$These authors contributed equally to this work.}%
\thanks{Digital Object Identifier (DOI): see top of this page.}
}


\maketitle

\begin{abstract}
While tactile sensor technology has made great strides over the past
decades, applications in robotic manipulation are limited by aspects
such as blind spots, difficult integration into hands, and low spatial
resolution. We present a method for localizing contact with high
accuracy over curved, three dimensional surfaces, with a low wire count
and reduced integration complexity. To achieve this, we build a volume
of soft material embedded with individual off-the-shelf pressure sensors. Using data driven techniques, we map the raw signals
from these pressure sensors to known surface locations and indentation depths. Additionally, we show that a finite element
model can be used to improve the placement of the pressure sensors
inside the volume and to explore the design space in simulation. We
validate our approach on physically implemented tactile domes which
achieve high contact localization accuracy ($1.1mm$ in the best case)
over a large, curved sensing area ($1,300mm^2$ hemisphere). We believe
this approach can be used to deploy tactile sensing capabilities over
three dimensional surfaces such as a robotic finger or palm.

\end{abstract}



\section{Introduction}

\IEEEPARstart{T}{actile} sensing has long been identified as a promising capability for
versatile robotic manipulation. As a result, the field has a rich
history including numerous transduction methods, sensor designs and
calibration methods, sensitivity analyses, etc.; details can be found
in extensive reviews covering the
field~\cite{DAHIYA10,HAMMOCK13,kappassov2015}. Even though
sensor technology has seen impressive advances, recent studies point
out that significant limitations still exist from the perspective of
applying this technology to robotic
manipulation~\cite{JENTOFT14,WAN16}. In particular, three identified
limitations of tactile sensors for robot manipulation include blind
spots, difficulty of hardware integration into palms and fingers, and
low spatial resolution.

Such limitations are particularly vexing because they are
intertwined. Improving spatial localization accuracy by adding more
sensors leads to increased wiring, which makes integration more
difficult. Integrated taxel arrays on PCBs reduce wiring, but robot
fingertips present many surfaces with complex 3D curvature; so planar
sensor arrays inevitably leave blind spots. Our work aims to simultaneously address all three limitations by achieving contact sensing with high localization accuracy over
non-flat three dimensional surfaces with reduced number of wires.

\begin{figure}[t]
\centering
\includegraphics[clip, trim=1.3cm 1cm 2.3cm 2cm,
  width=0.90\linewidth]{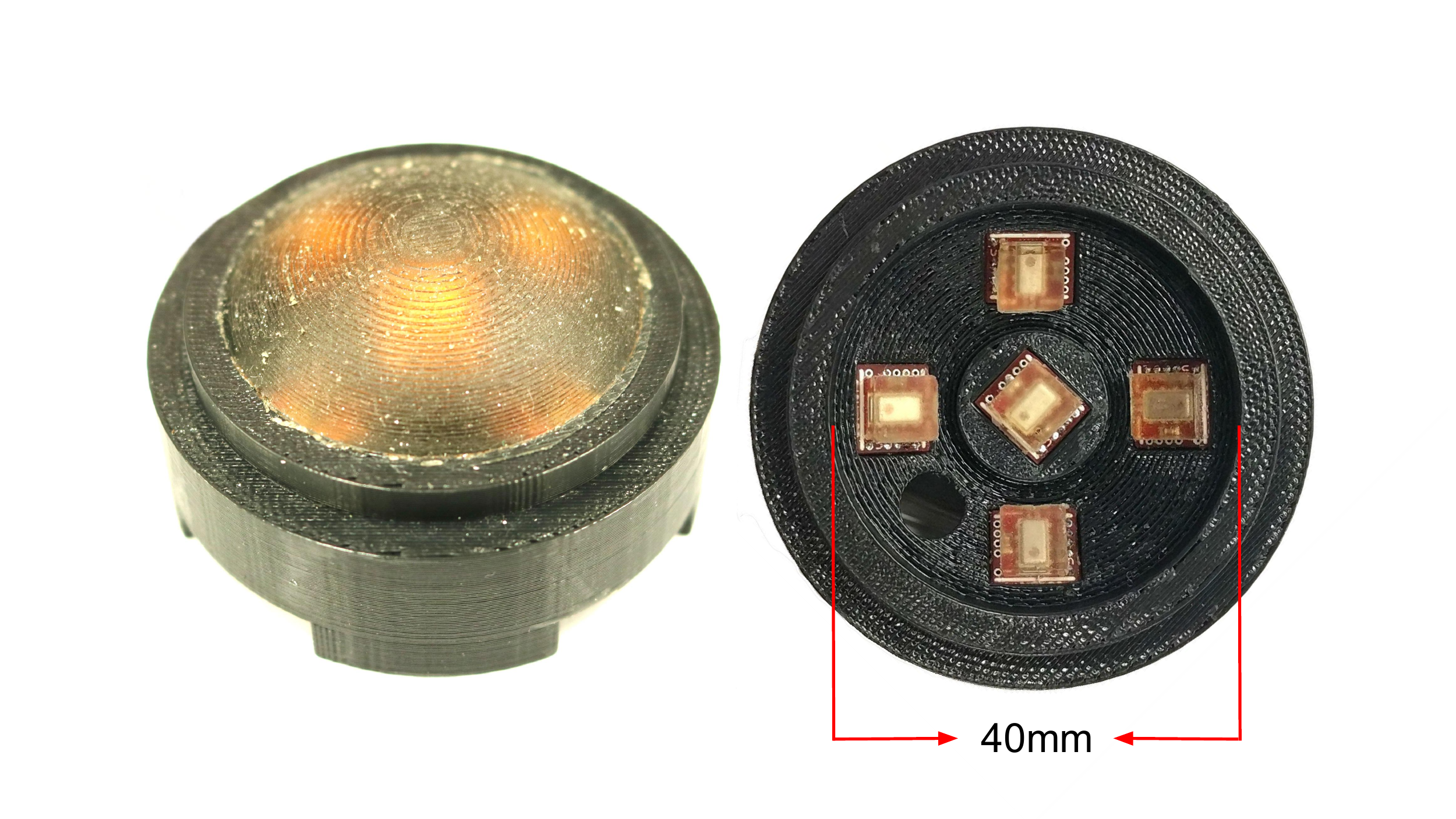}
\caption{Example of a soft volume embedded with individual pressure sensors. This tactile dome has a hemispherical surface and
  five embedded Takktile pressure sensors. We use data driven
  techniques to localize touch across its three dimensional surface.}
\label{fig:tactile_dome}
\end{figure}

Our method is to fabricate a volume of soft material embedded with just a few off-the-shelf, individual pressure sensors. We demonstrate this approach on a hemispherical body, a ``tactile dome'', with multiple Takktile pressure sensors~\cite{TENZER14} (Figure~\ref{fig:tactile_dome}). Our objective is to localize touch on this three dimensional
surface and to find what sensor configurations within
the volume improve localization accuracy.

Our key intuition is that sensor configurations that favor cross-talk
will provide better localization: if an indentation anywhere on the
dome's surface produces a meaningful signal in as many sensors as
possible, such a ``signature'' can be used to determine the location
of the indentation. This can be considered an instance of
super-resolution: we are trying to engineer overlapping receptive
fields for each Takktile pressure sensor such that we can perceive
stimuli in finer detail than if we treat each sensor as an individual
taxel. 

What distinguishes our approach from traditional super-resolution
methods is that we allow any configuration of pressure sensors inside
the volume. We use a purely data-driven method to find the mapping
between the signals reported by each Takktile sensor and the location
of the indentation. We record data from the pressure sensors while
indenting the dome at known locations and to a certain depth. We then
train a machine learning algorithm based on this data to predict the
indentation location based on sensor readings.

A purely data-driven mapping method (with no analytical model) opens the question of how to place the sensors inside the soft volume
in order to achieve the desired localization accuracy. To address
this, we implement a quasi-static finite element analysis (FEA)
simulation which we show is a reliable predictor for the behavior
of real sensors. The FEA simulation generates synthetic datasets which
we feed into our machine learning algorithm to predict the
performance of a given sensor configuration. This enables us to test different configurations virtually without having to build a physical
sensor. We demonstrate this by testing ten different sensor
configurations in simulation. Moreover, we build two of the simulated models into physical tactile domes for comparison and validation.

The main contributions of this paper can be summarized as follows: 
\begin{itemize}
\item We present a soft hemispherical sensor capable of localizing
  touch over its three dimensional surface with high accuracy ($1.1mm$
  median error in the best case).
\item Our tactile dome uses relatively few embedded sensors (five), along with a data-driven localization method, to
  provide coverage over a large area ($1300mm^2$). This reduces the
  overall cost of the sensor, minimizes the wiring, and simplifies
  future integration into hands.
\item We use finite element analysis to improve sensor performance by
  iterating over the design parameters in simulation. This eliminates 
  the need to physically build a new sensor. We validate the simulation against
  real implementations of multiple Takktile sensor configurations and show how sensor
  placement affects individual pressure trends as well as overall
  performance.
\end{itemize}
To the best of our knowledge, we are the first to show that a purely
data-driven method can use relatively few embedded sensors to achieve
high accuracy contact localization over a large three dimensional
surface. We are also the first to explore the relationship between
pressure sensor distribution inside a three dimensional volume,
cross-talk between the sensors induced by an indentation on the
surface, and the indentation localization accuracy that results. We
believe this method can be used to develop tactile sensing
capabilities over three dimensional surfaces, enabling better coverage
over robotic fingers and hands.

\section{Related Work}

When building tactile sensors, numerous transduction methodologies have been explored. We refer the reader to a number of comprehensive reviews~\cite{DAHIYA10,HAMMOCK13,kappassov2015}. The work presented here does not intend to explore a new sensing modality; but to expand upon such modalities and find a method to deploy them over three dimensional surfaces.

We use Takktile from Right Hand Robotics~\cite{TENZER14} as the sensing elements in our tactile dome. While these sensors are mainly designed to measure normal forces, Reeks et al.~\cite{reeks2016angled} has shown that positioning these sensors at an angle can produce meaningful signals to measure shear forces. In contrast, our work focuses on contact localization over a three dimensional surface and does not provide force measurements.

Using eight Takktile sensors, Guggenheim et al.~\cite{guggenheim2017} built a 6 axis force-torque sensor by arranging the sensors in a particular spatial configuration, and fitting an analytical model to the calibration data. This work shows that appropriate spatial arrangement of Takktile sensors can be used to measure multi-axial forces. Although our work here solely focuses on touch localization, we show that the concept of optimizing the geometry and positioning of the pressure sensors inside a volume of soft material can yield significant improvements in the overall sensor performance. 

A similar sensor to our tactile dome was presented by Chathuranga et al.~\cite{chathuranga2016soft} using a permanent magnet embedded in a hemispherical volume of silicon rubber. Three orthogonal hall effect sensors are embedded into the rubber and their individual measurements allow them to determine the displacement of the magnet, which is used to calculate the applied force. Another analogous magnetic-based sensor was built by Paulino et al.~\cite{paulino2017} using a single Hall-effect based tri-axis magnetometer to measure the displacement of the permanent magnet embedded in the elastomer. They improved sensitivity by intentionally creating an air gap between the elastomer and the magnetometer. Once again both of these sensors focus on determining the applied forces but do not provide touch localization capabilities. 

Our localization approach can be thought of as an application of super-resolution techniques. In general, super-resolution for tactile sensing leverages overlapping receptive fields of neighbouring taxels to perceive stimuli detail finer than the sensor resolution. Van den Heever et al.~\cite{HEEVER09} used several measurements of a 5 by 5 force sensitive resistor array and combined them into an overall higher resolution measurement. Lepora and Ward-Cherrier~\cite{LEPORA151} and Lepora et al~\cite{LEPORA152} used a Bayesian perception method to obtain a 35-fold improvement of localization acuity ($0.12mm$) over a sensor resolution of $4mm$. In our design it is difficult to establish an initial resolution metric, but the super-resolution concept to leverage overlapping receptive fields is exploited to provide contact localization. We are intentionally exploring how to place the pressure sensors inside the volume of material such that any indentation produces a signal on several sensors. 

We use a data-driven method to train our sensor and learn the mapping from the five Takktile sensors to the location of the indentation. We have used a similar approach with success in the past when building planar tactile sensors~\cite{PIACENZA16_IROS, PIACENZA17_ICRA}. Along these lines we note that the use of machine learning methods for manipulation based on tactile data is not new. Ponce Wong et al.~\cite{PONCE14} used signals provided by a previously developed~\cite{WETTELS08} multimodal touch sensor to discriminate between different geometric features. Dang and Allen~\cite{ALLEN13} were able to successfully differentiate stable versus unstable grasps using a support vector machine (SVM) classifier in the context of robotic manipulation using a Barrett Hand. Bekiroglu et al.~\cite{BEKIROGLU11} studied how grasp stability can be assessed based on tactile sensory data using machine-learning techniques. In similar fashion, both Saal et al. \cite{SAAL10} and Tanaka et al.~\cite{TANAKA14} used probabilistic models on tactile data to estimate object dynamics and perform object recognition respectively. However, most of this work is based on arrays built on rigid substrates and thus unable to provide full coverage of complex geometry. In contrast, we apply our methods to the design of the tactile dome itself, and believe that developing the sensor simultaneously with the learning techniques that make use of the data can bring us closer to achieving complete tactile systems.

\section{Tactile Dome Design and Data Collection}

\begin{figure}[t]
\includegraphics[clip, trim=1.0cm 1.0cm 1.0cm 1.0cm, width=0.95\columnwidth]{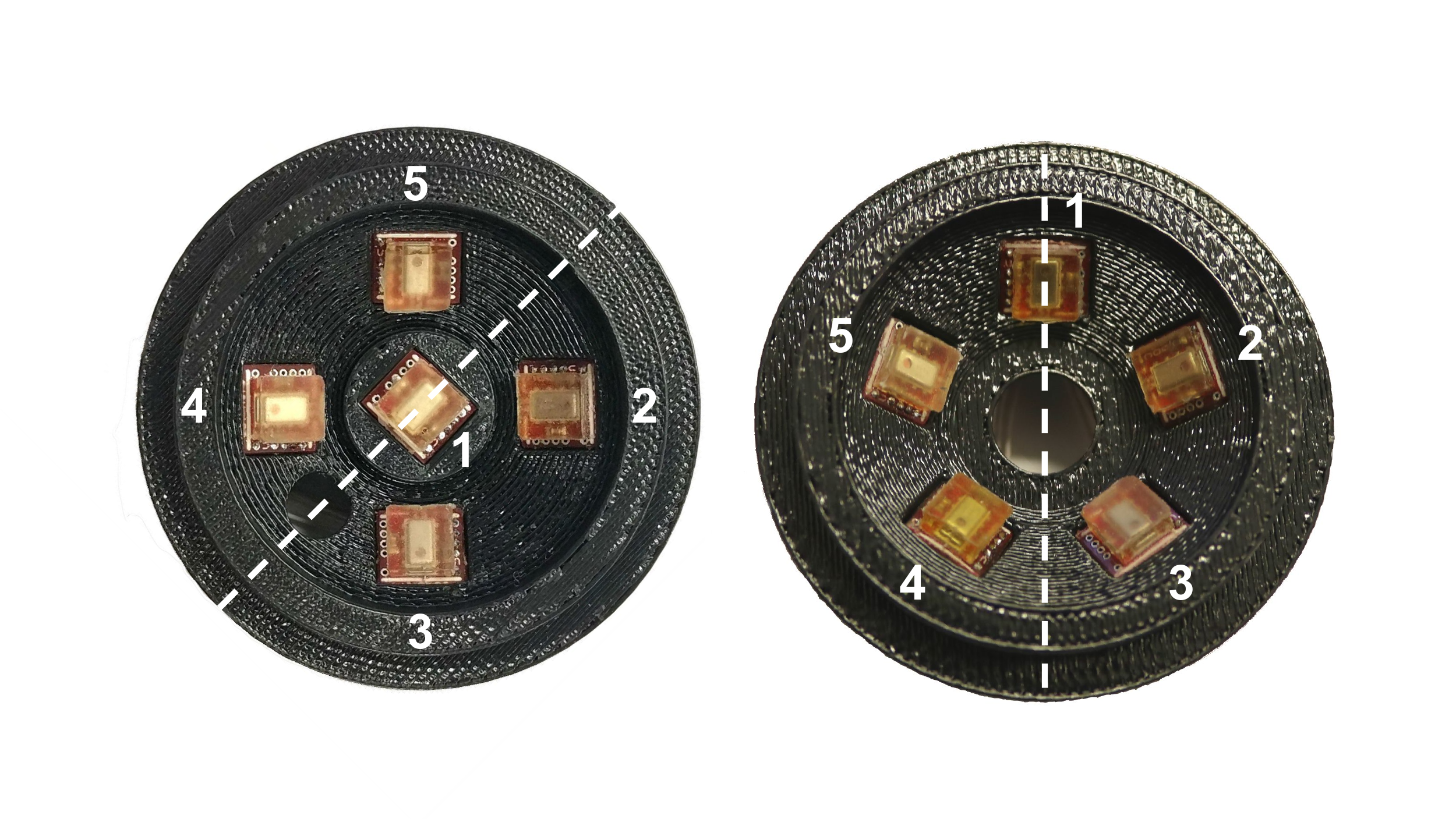}
\caption{Top view of two distinct sensor configurations inside a tactile dome, each showing a dotted symmetry line and numbering for individual sensors.}
\label{fig:distributions}
\end{figure}

The tactile dome is designed to represent a generic, three dimensional surface such as an anthropomorphic fingertip or palm. A 3D-printed base made of ABS material acts as a mounting structure for both the soft volume and the Takktile sensors. A second piece mates with this base to create the casting mold with a spherical surface. The useful area of the sensor is the section of a $54mm$ diameter sphere subtended by a cone with a 90 degree aperture, resulting in a surface of approximately $1300mm^2$. This design is illustrated in Figure~\ref{fig:tactile_dome}.

The tactile dome is cast on the base using Vytaflex 20, the same rubber urethane used on the Takktile sensors to create a cohesive bond. Prior to casting, the solution is degassed to ensure that the tactile dome has consistent, isotropic properties. The two constructed physical sensors have a hole in the ABS base where the solution is poured while the assembly is upside down. The solution is cured for 24 hours at room temperature and then placed in an oven at 65 C for 8 hours. The hole used to pour the solution is then plugged using a matching ABS piece and epoxy. 

We aim to find a configuration that achieves high accuracy for touch localization using only five Takktile sensors. To this end, we explore 10 possible sensor configurations, using both simulation and fabrication of real sensors. We will discuss all of these configurations in detail in the following sections. First we introduce our data collection and analysis protocols below, which are the same regardless of the sensor configuration.

\subsection{Data Collection Methodology}

\begin{figure}[t]
\centering
\includegraphics[clip, trim=2cm 2cm 2cm 2cm,
  width=0.90\linewidth]{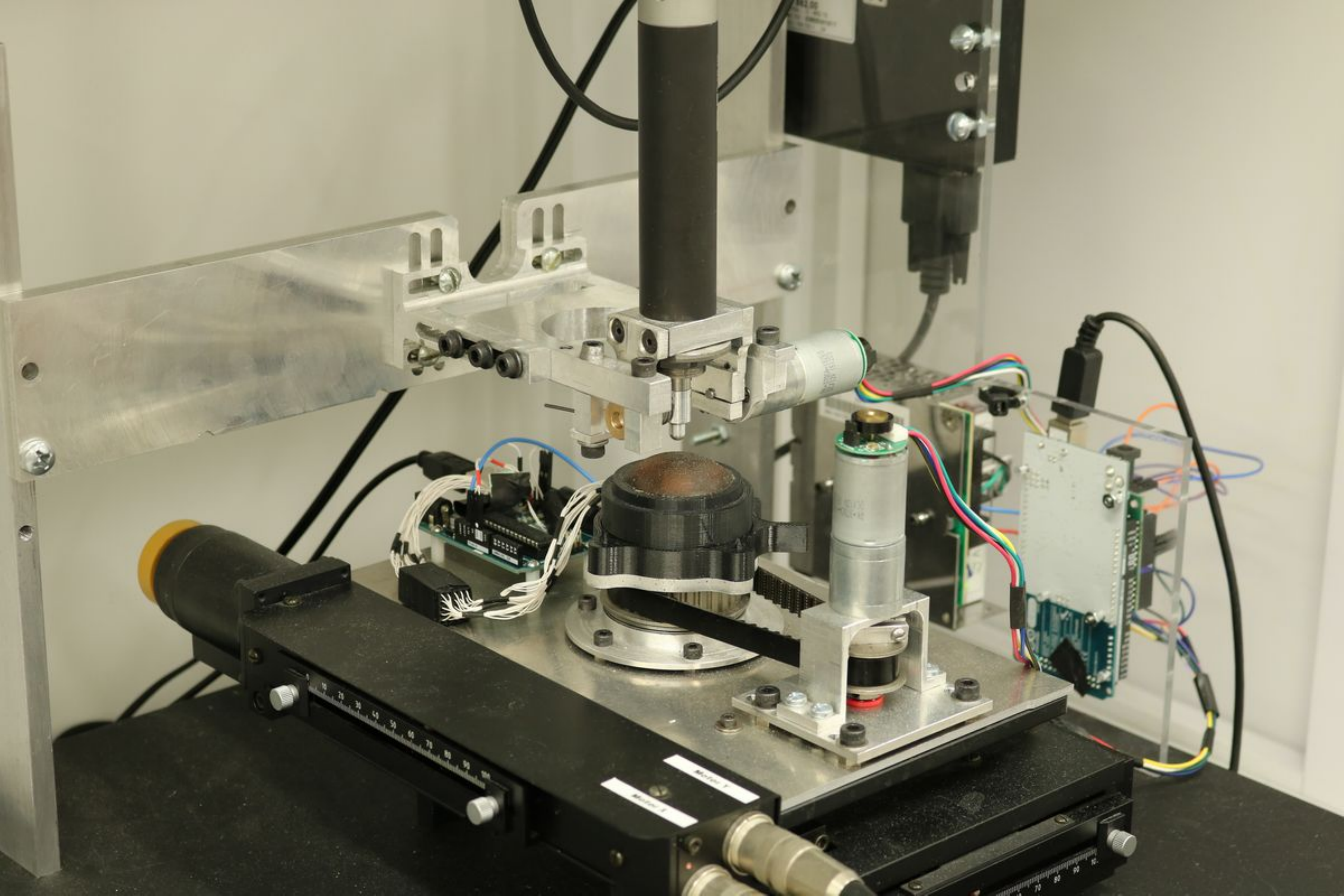}
\caption{Our indenter machine is composed of a planar stage and a linear actuator. The tactile dome can rotate on top of the planar stage and the linear actuator can pivot up to 60 degrees on an axis parallel to the stage. These additional degrees of freedom were built such that the tactile dome can be indented at any point with any angle with respect to the surface normal.}
\label{fig:indenter_machine}
\end{figure}

Our goal is to find a data-driven mapping between the five pressure sensor readings and the location of an indentation. To collect training data, we use an indenter machine with five degrees of freedom (DOF) capable of probing at specified locations and depths along the dome's surface with varying angles from the surface normal (Figure~\ref{fig:indenter_machine}). The indenter machine is composed of a planar stage (Marzhauser LStep) where we attach our tactile dome on top of a rotating base. We mount a position controlled linear actuator (Physik Instrumente M-235-5DD) on a revolute joint directly above the dome. The linear actuator is equipped with a $6mm$ diameter aluminum hemispherical tip. The only unaccounted DOF is a revolute joint around the tool tip axis. In this study all indentations are performed with the linear actuator positioned normal to the dome's surface. 

To collect training data, we sample the tactile dome at equidistant locations along its surface. To achieve this uniform distribution over a spherical surface, we use a mapping scheme \cite{rocsca2010new} to project a two dimensional regular grid $(A,B)$ onto the dome's hemispherical surface $(x,y,z)$. Using $A,B \in [-15,15]$ results in a mapping that almost completely covers our three dimensional surface. Note that this $(A,B)$ space is dimensionless. We use a 16x16 regular grid, hence we sample along 256 different positions on the surface of the dome for training. At each location, the probe is positioned normal to the surface and a first ``non-touch'' data point is recorded before the linear actuator tip contacts the dome's surface. Subsequently, the actuator indents the dome and records data at the predefined depths from $0$ to $3mm$ in $0.5mm$ intervals. 

To collect a test dataset, we follow an identical procedure except indentation locations are randomly sampled from our two dimensional $(A,B)$ space. We collect data on 100 random locations. The test dataset locations, while generated randomly within our $(A,B)$ space, are repeated for all 10 sensor configuration cases (both real and simulated) for consistency.  

For each measurement $i$, we record the data in a tuple of the form $\Phi_i = (A,B,d_i,r_i^1,....,r_i^5)$ where $(A,B)$ is the location of the indentation before being mapped to the dome's surface, $d_i$ is the depth of the measurement, and $r_i^1,....,r_i^5$ are the readings of the five pressure sensors. These tuples are used as described in the data-driven localization algorithm section. 

\subsection{Simulation}

In addition to the methodology described above for building and using physical sensors, we have developed a corresponding simulation framework based on finite element analysis (FEA). The purpose of developing a simulated model of our tactile dome is to have the ability to iterate and evaluate sensor configurations to improve the localization performance without the need to build the physical tactile dome. 

In the simulated case, we use a quasi-static, ABAQUS Standard FEA to generate both training and testing data. The indentations are performed with a $6mm$ diameter, hemispherical, rigid body at the same locations as on the physical tactile dome, normal to its surface. The indenter interacts with the tactile dome's surface with an estimated friction coefficient of 0.8. We perform a uniaxial tensile test to achieve accurate material properties for the specific batch of Vytaflex 20 used to fabricate the physical dome. This experimental stress-strain data is input into ABAQUS and the software's Marlow curve fitting model predicts the material properties.

To simplify the model, we make the following assumptions:
\begin{itemize}
\item The tactile dome's polymer is perfectly bonded to all of the components it interfaces with during casting.
\item The compliances of the ABS base, Takktile sensor housing, and Takktile circuit are ignored. These components are treated as infinitely stiff compared to the polymer.
\item The Takktile sensor samples pressure at its air-vent hole, which is an opening to the barometer's pressure-sensing Wheatstone bridge. The elastomer is assumed to have the same properties as Vytaflex 20.
\end{itemize}

The first two assumptions enable us to exclude the ABS base, Takktile sensor housing, and Takktile circuit from the simulation in order to reduce computational time and complexity. To compensate, a zero-displacement constraint in the X-Y-Z directions is applied on the surfaces where the polymer interacts with these components. It should be noted that these surfaces are not fixed in rotation to ensure that the simulation is not over-constrained. 

To reproduce the interaction between the tactile dome and the Takktile sensor air-vent hole, five thin, cylindrical bodies are attached to the dome via a ``tie'' constraint, representing perfect bonding of Vytaflex 20. The opposing cylindrical faces are fixed in the X-Y-Z directions to mimic the attachment to the Takktile sensor's internal strain gauge. The cylinders allow us to sample at a location that is not a boundary condition such that we generate a realistic output. We sample pressure data at the surface between the cylindrical bodies and the tactile dome.  

In order to shorten the computational time required for the solution to converge, data is collected over half of the sensor and mirrored about the line of symmetry. The symmetry line corresponds in $(A,B)$ space to the line A=-B for configurations with a central sensor and to $A=0$ for configurations with all sensors arranged in a circumference (see Figure~\ref{fig:distributions}).

\begin{figure}[t]
\setlength{\tabcolsep}{0mm}
\begin{tabular}{cc}
\subfloat[C3D4H mesh] {\label{mesh_4}
\includegraphics[clip, trim=0.4cm 0.5cm 0.5cm 0.3cm, width=0.49\columnwidth]{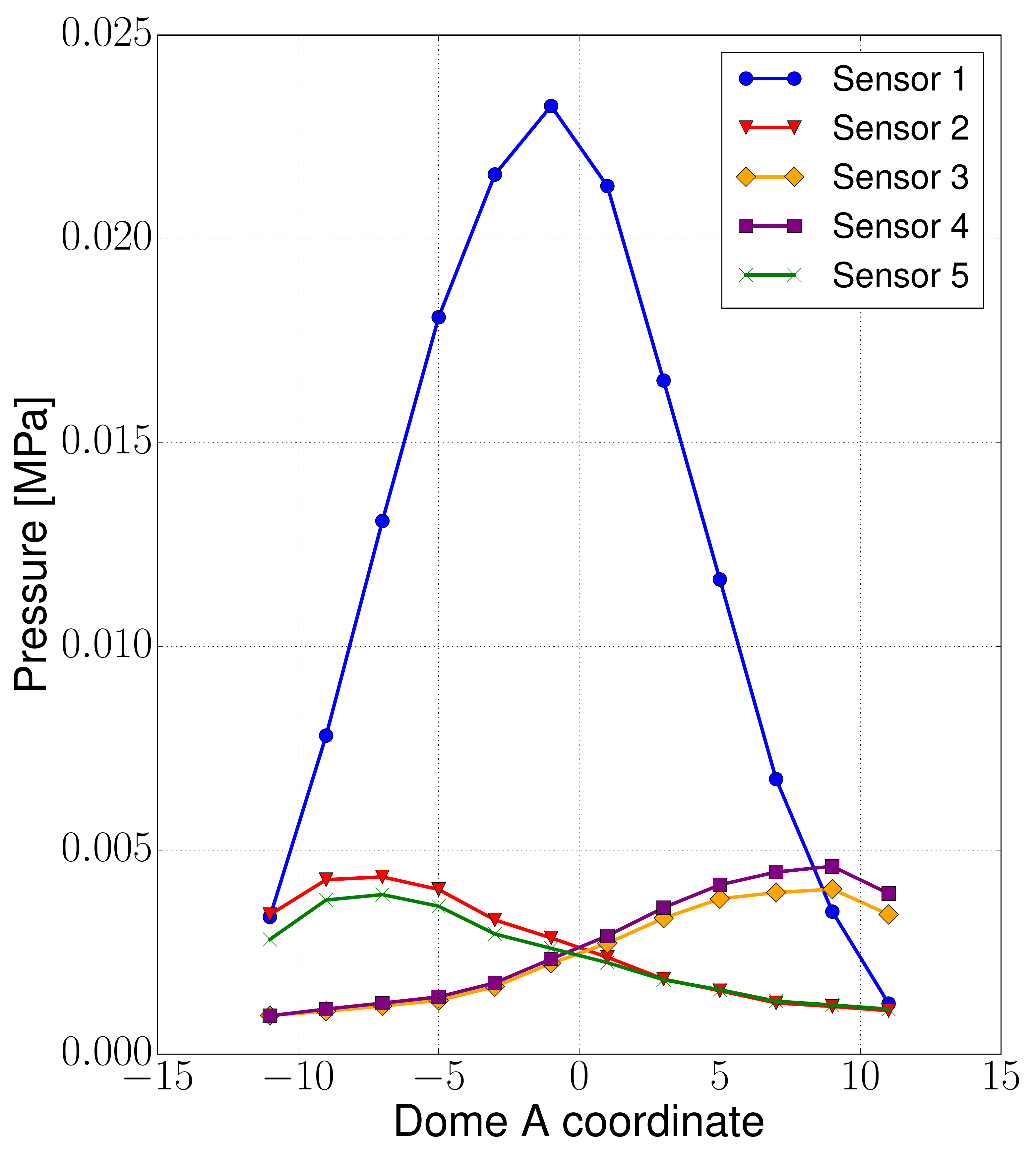}}\hfill &  \subfloat[C10D4i mesh] {\label{mesh_10}
\includegraphics[clip, trim=0.4cm 0.5cm 0.5cm 0.3cm, width=0.49\columnwidth]{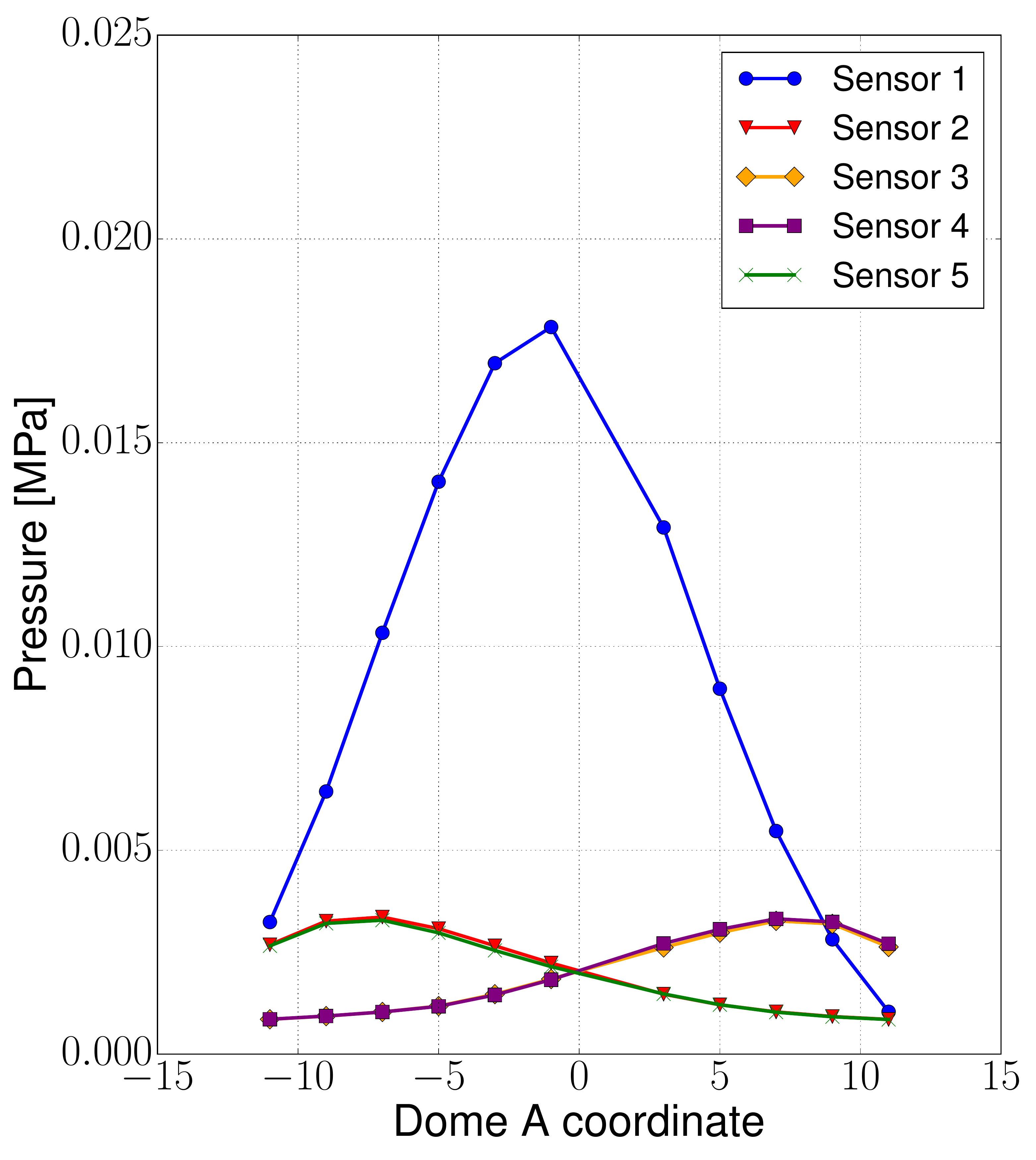}}\\
\end{tabular}
\caption{Comparison between simulations ran using a C3D4H mesh (\ref{mesh_4}) and a C10D4i mesh (\ref{mesh_10}) on Case 1. The graph represents the pressure distribution among the five Takktile sensors as we indent along the symmetry line. Both graphs show similar trends with slightly different absolute values for the reported pressure.}
\label{fig:mesh_study}
\end{figure}

The dome and five bodies are modeled with a first-order mesh with hybrid elements due to the incompressible properties of Vytaflex 20 and the curved geometry. We ran a mesh convergence study using Case 1 and determined that a 10-node, C10D4i mesh yields similar trends as a 4-node C3D4H mesh (Figure~\ref{fig:mesh_study}). Additionally, the average time to simulate a single indentation using the C10D4i mesh is more than 10 times longer than using a C3D4H mesh; therefore, we use the 4-node mesh in order to decrease the simulation run time.

\section{Data-driven Localization Algorithm}

The goal of our localization algorithm is to learn the mapping from
Takktile pressure sensor signals $(r^1,...,r^5) $ to the location of
the indentation $(A,B)$. For a given sensor configuration, we train a
regressor to learn this mapping using a training dataset containing
256 datapoints, each corresponding to one location in the $(A,B)$
grid. We then test performance on a test dataset, containing data from
indentations at 100 random locations. For consistency, we use the same
test set locations for all sensor configurations.

We use this procedure in exactly the same way on both physical and
simulated domes. Real and synthetic data are treated equally, except
for the fact that simulated training data is mirrored about the
symmetry axis (to save simulation time), whereas real data is
collected over the entirety of the $(A,B)$ grid.

The predictor we use is a kernelized ridge regressor with a laplacian
kernel. We have found that the laplacian kernel provides better
results compared to the radial basis function kernel (comparison data
not included here), and thus used the laplacian kernel for all results
shown in this paper. We also perform cross-validation using the k-fold
method with $k=5$ and shuffling the data on each fold. The regressor
hyper-parameters (the ridge regressor regularization term $\alpha$ and
the laplacian kernel parameter $\gamma$) are chosen through an
exhaustive grid search of 400 combinations: 20 values in logarithmic
space for both $\alpha$ and $\gamma$ ranging from $10^{-5}$ to
$10^{1}$. The metric used to decide on the best hyper-parameter
combination is to minimize the median error of the localization
prediction. All results are obtained using the scikit-learn Python
package implementation of the kernelized ridge regressor.

\begin{figure*}[t]
\setlength{\tabcolsep}{1mm}
\begin{tabular}{ccccc}
\includegraphics[clip, trim=5.5cm 3.5cm 5.5cm 4.5cm, width=0.18\textwidth]{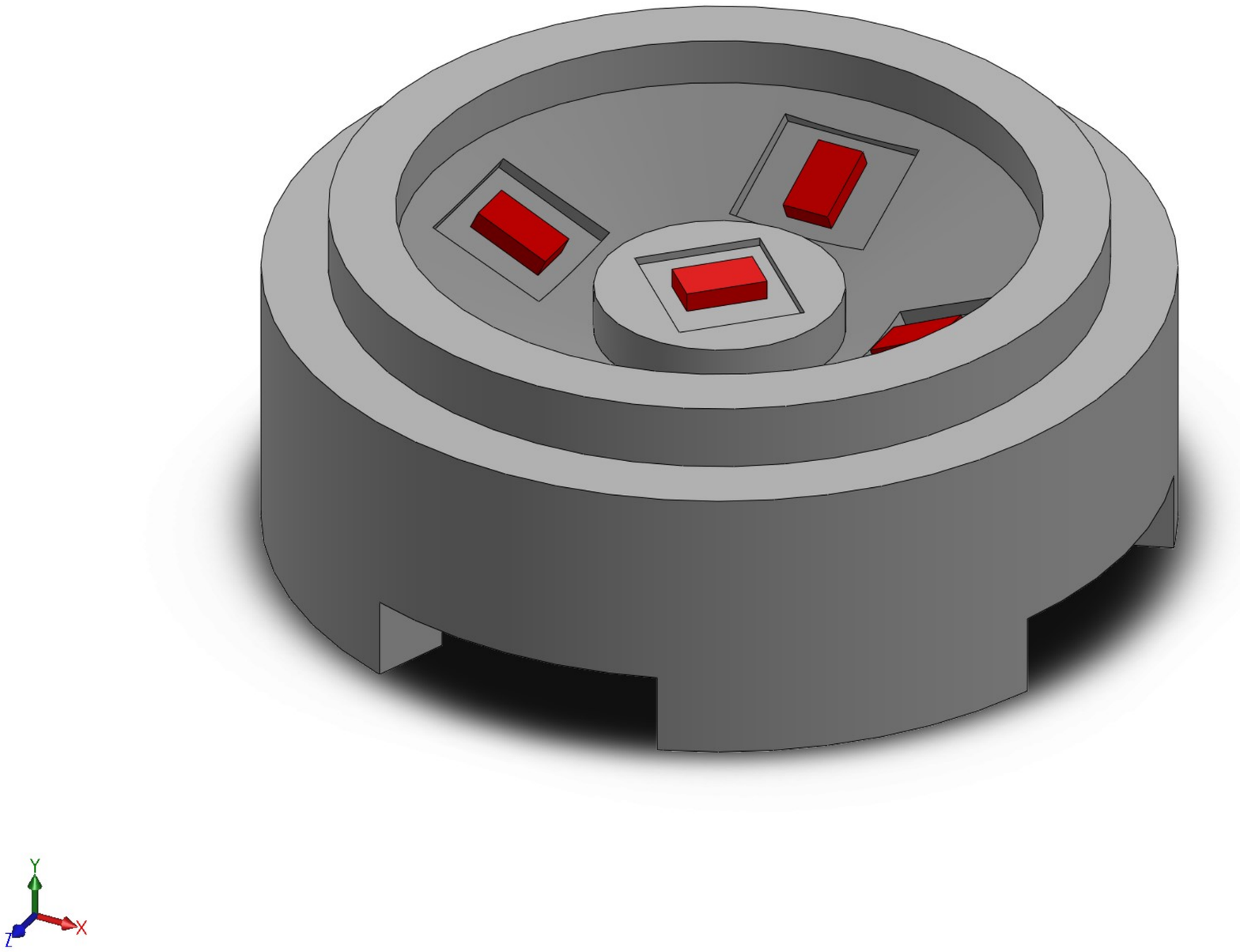}&
\includegraphics[clip, trim=4.9cm 3.5cm 4.9cm 3.9cm, width=0.18\textwidth]{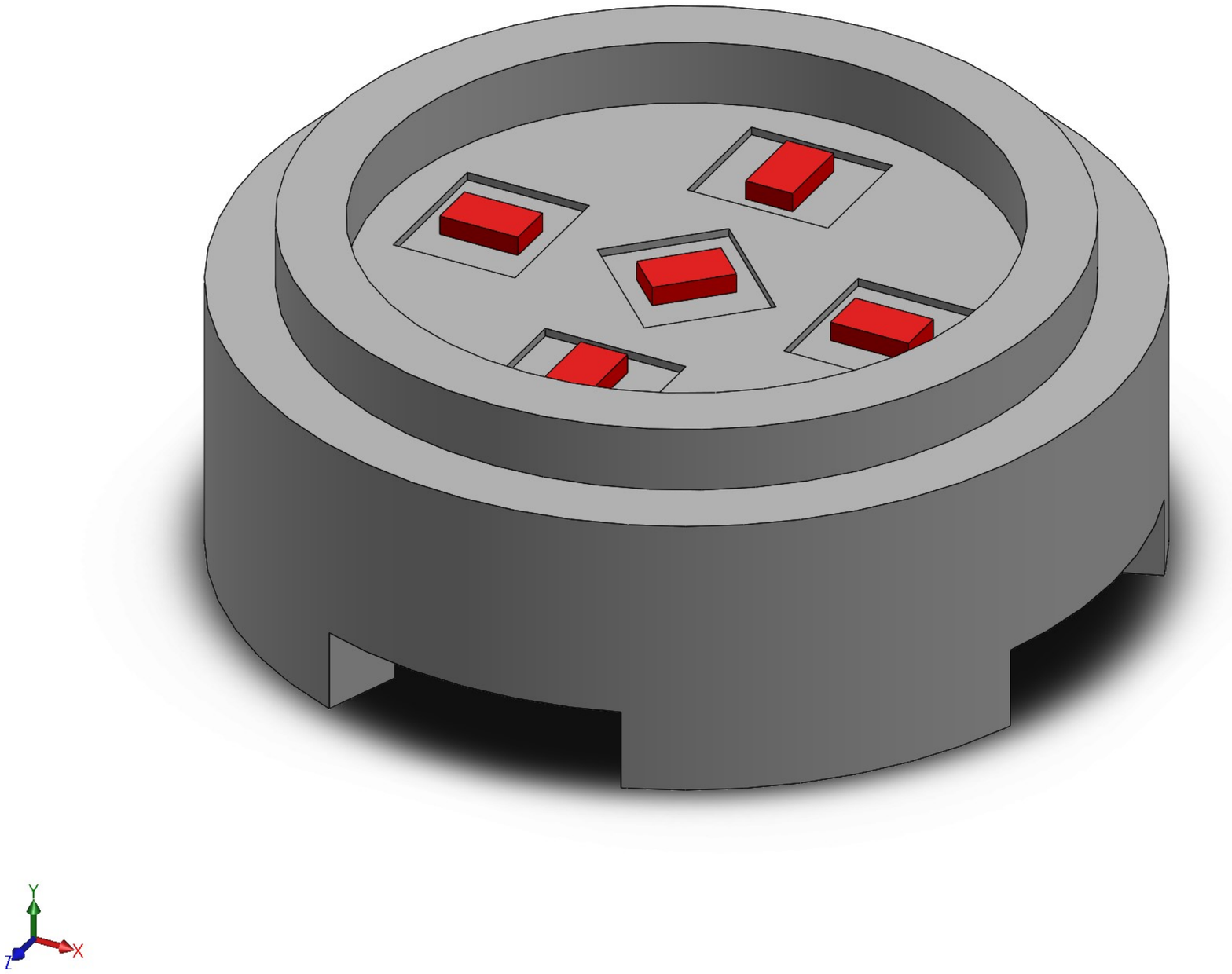}&
\includegraphics[clip, trim=4.2cm 2.8cm 4.2cm 3.8cm, width=0.18\textwidth]{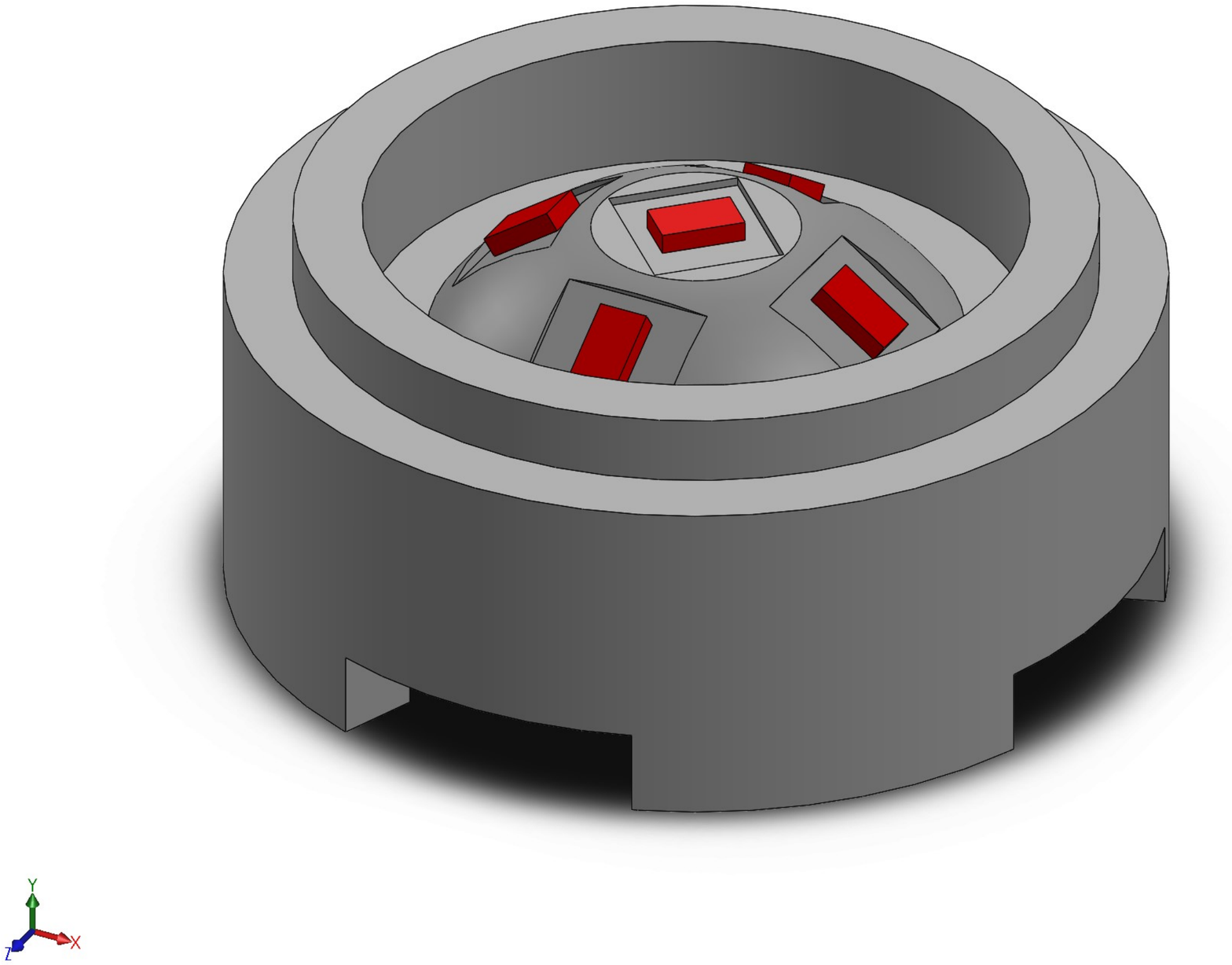}&
\includegraphics[clip, trim=4.9cm 3.5cm 4.9cm 3.9cm, width=0.18\textwidth]{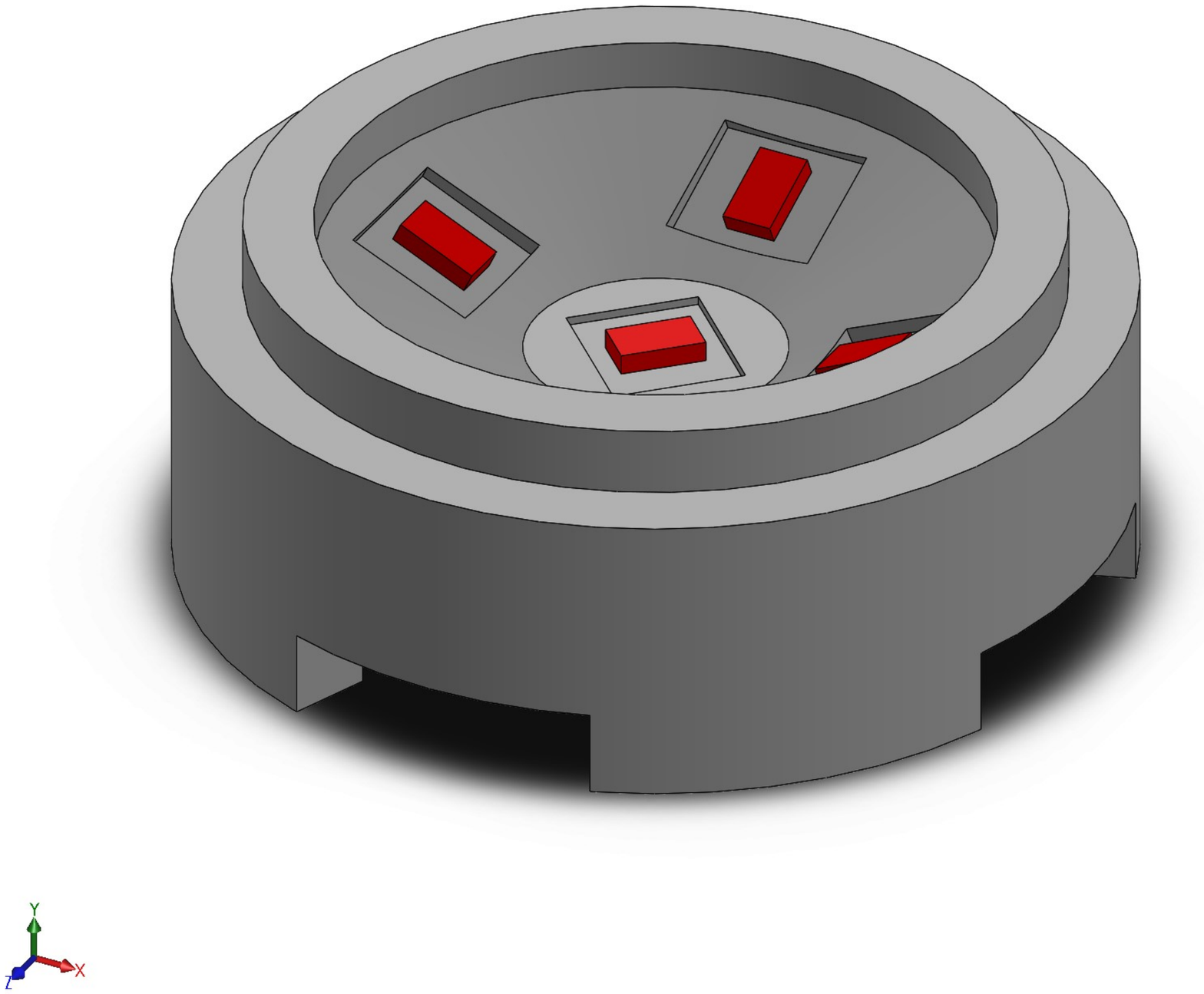}&
\includegraphics[clip, trim=5.5cm 3.5cm 5.5cm 4.5cm, width=0.18\textwidth]{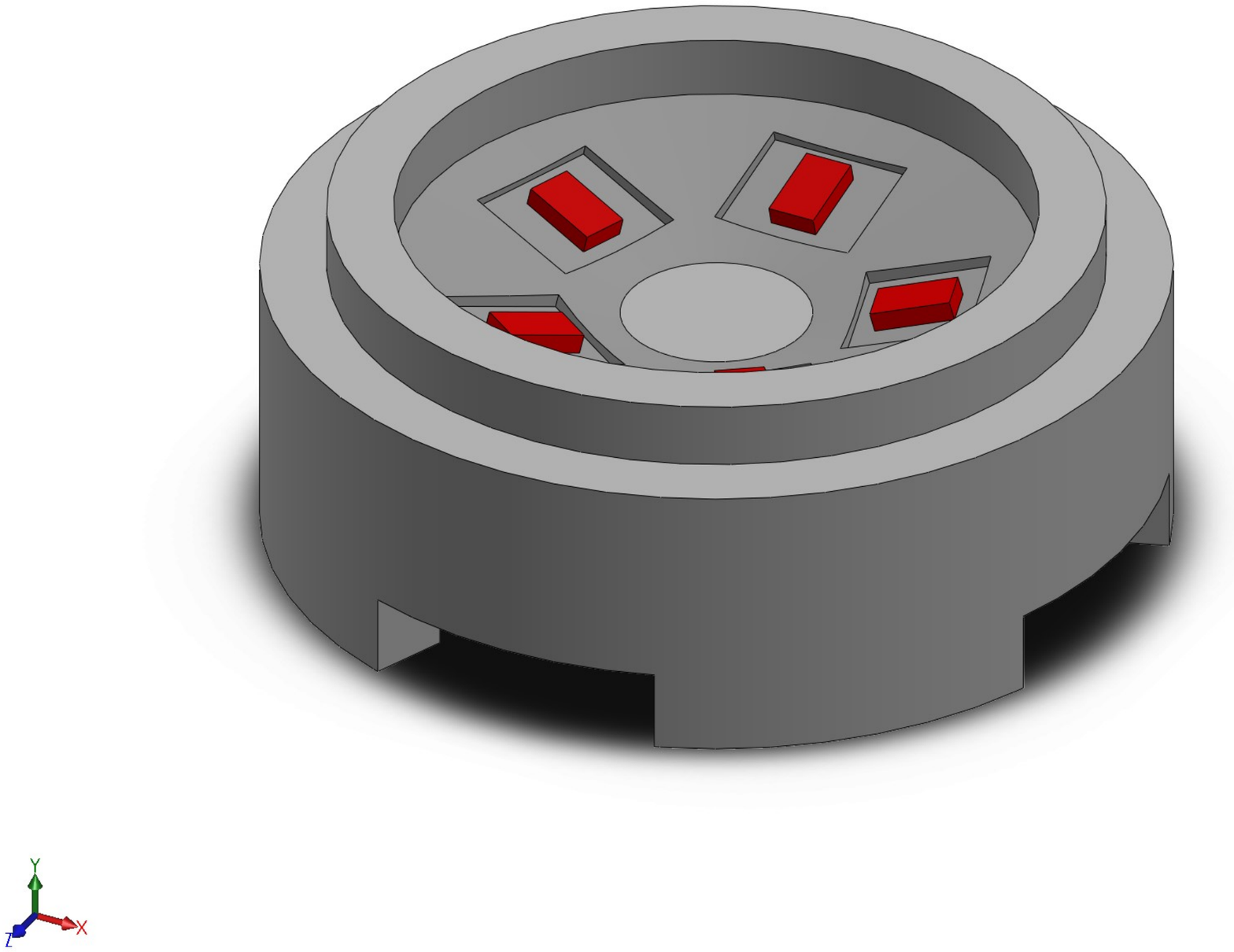}\\[0mm]
\footnotesize Case 1: Central 25$^\circ$  & 
\footnotesize Case 2: Central 0$^\circ$ & 
\footnotesize Case 3: Central -25$^\circ$ &
\footnotesize Case 4: Central 25$^\circ$ &
\footnotesize Case 5: Ring 15$^\circ$\\[-1mm]
\footnotesize with platform & & & \footnotesize without platform & \\[0mm] 

\includegraphics[clip, trim=5.5cm 3.5cm 5.5cm 4.5cm, width=0.18\textwidth]{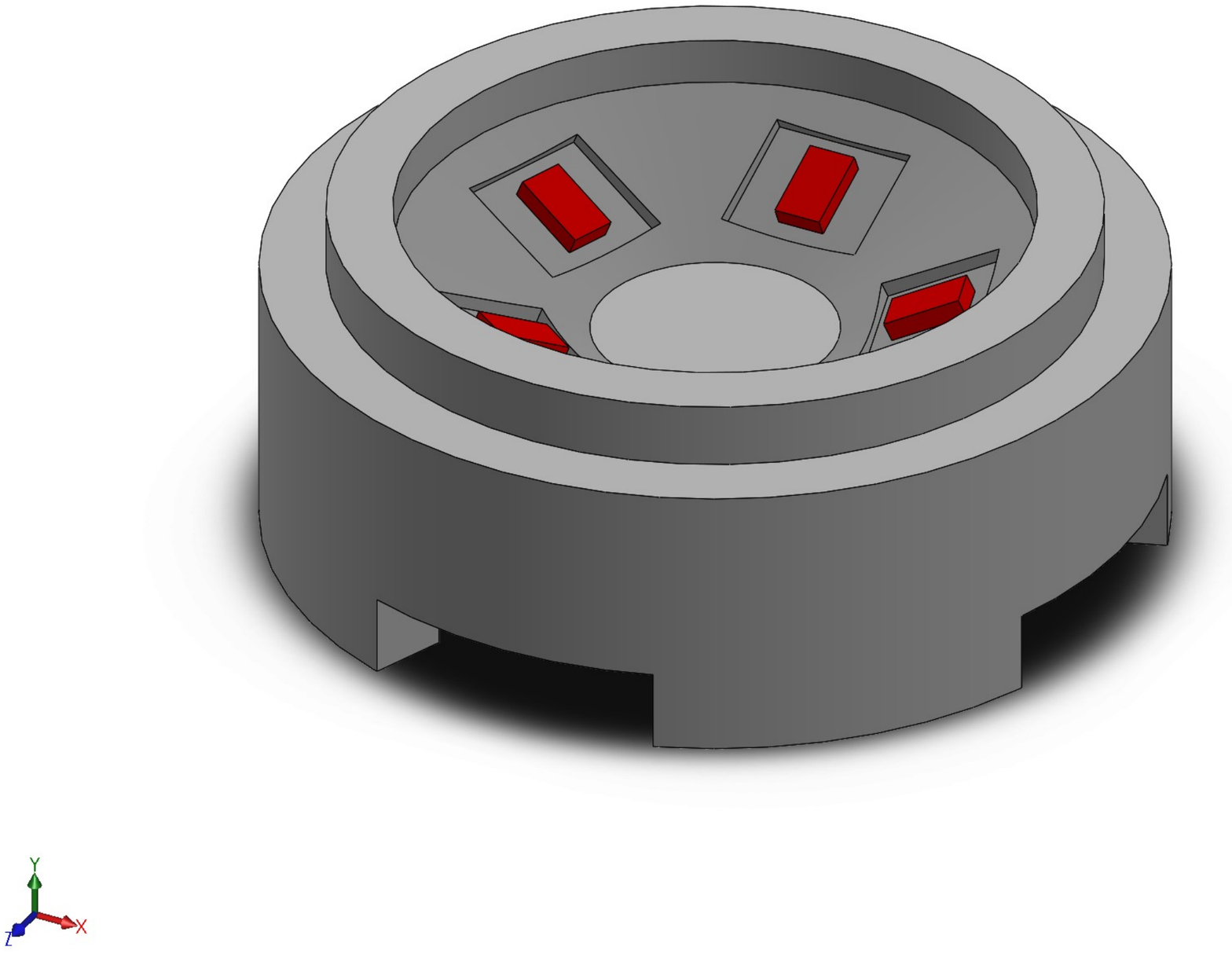}&
\includegraphics[clip, trim=5.5cm 3.5cm 5.5cm 4.5cm, width=0.18\textwidth]{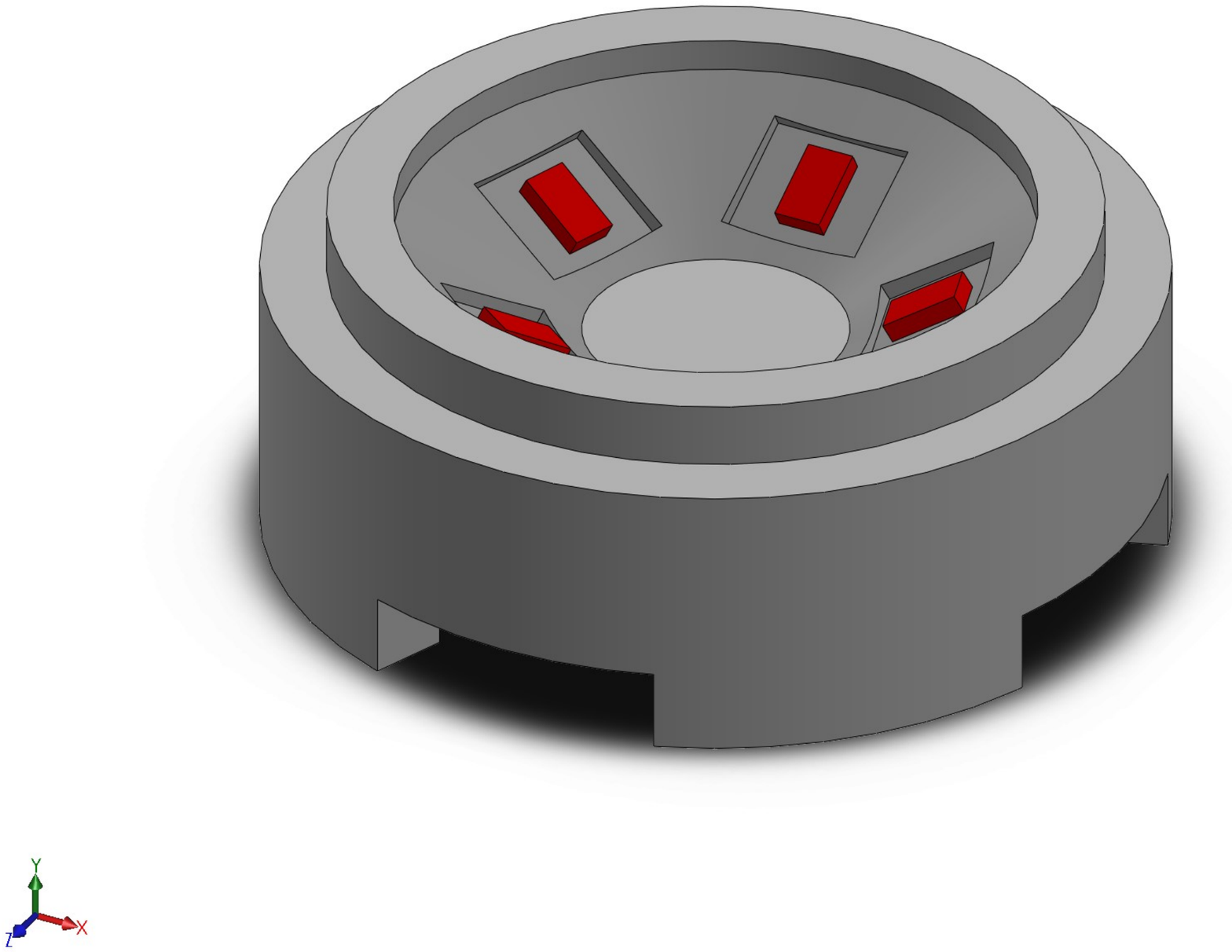}&
\includegraphics[clip, trim=5.5cm 3.5cm 5.5cm 4.5cm, width=0.18\textwidth]{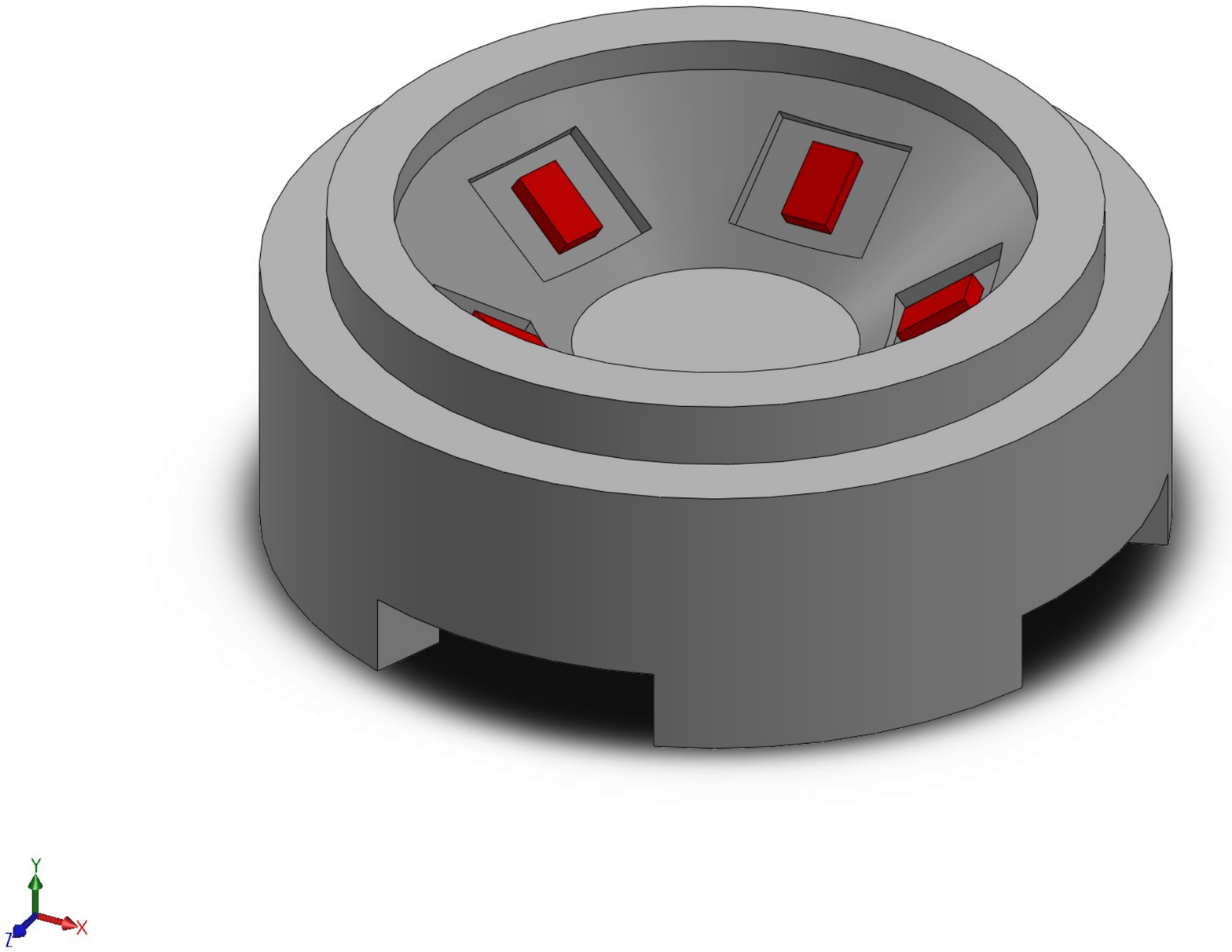}& 
\includegraphics[clip, trim=5.5cm 3.5cm 5.5cm 4.5cm, width=0.18\textwidth]{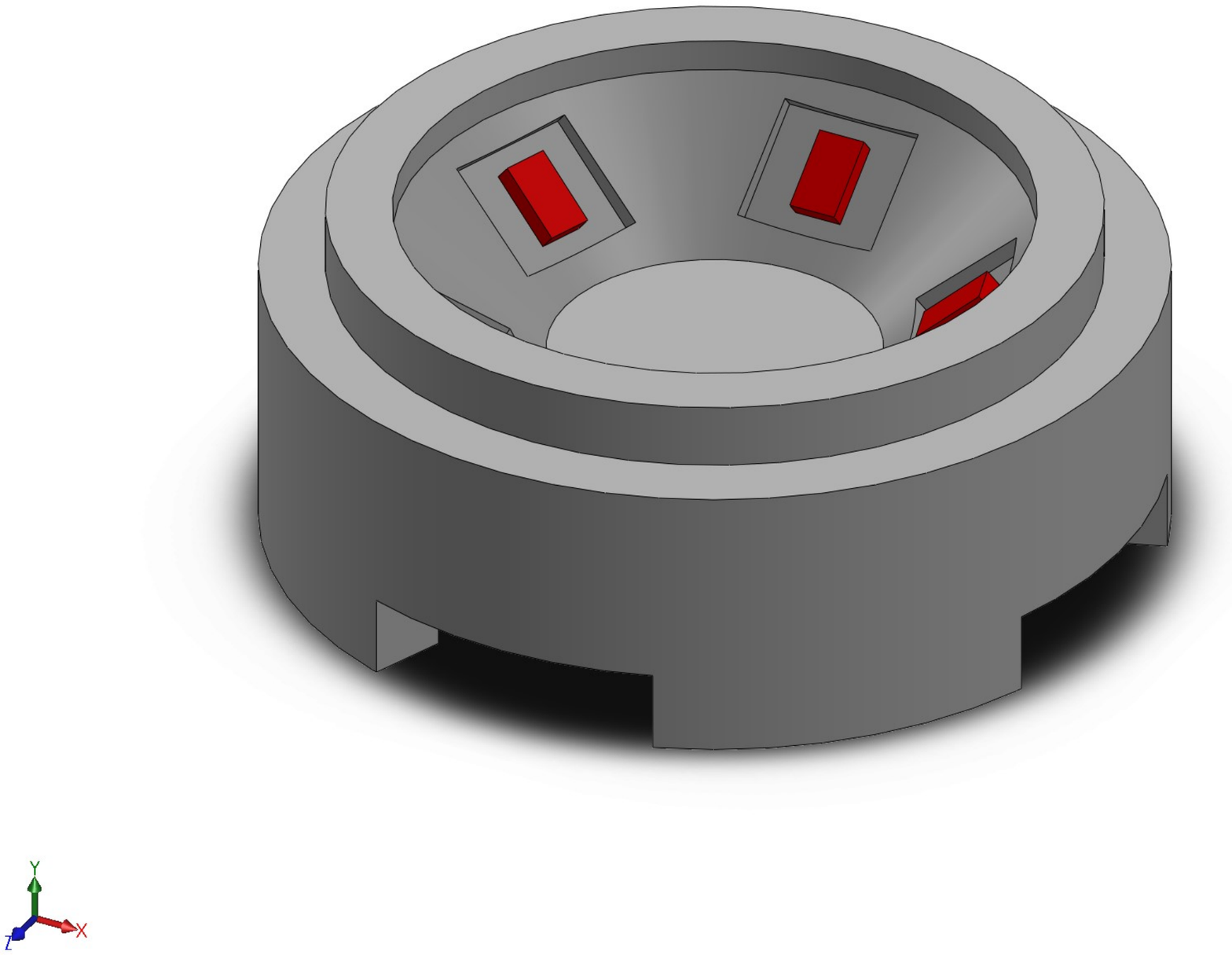}&
\includegraphics[clip, trim=5.5cm 3.5cm 5.5cm 4.5cm, width=0.18\textwidth]{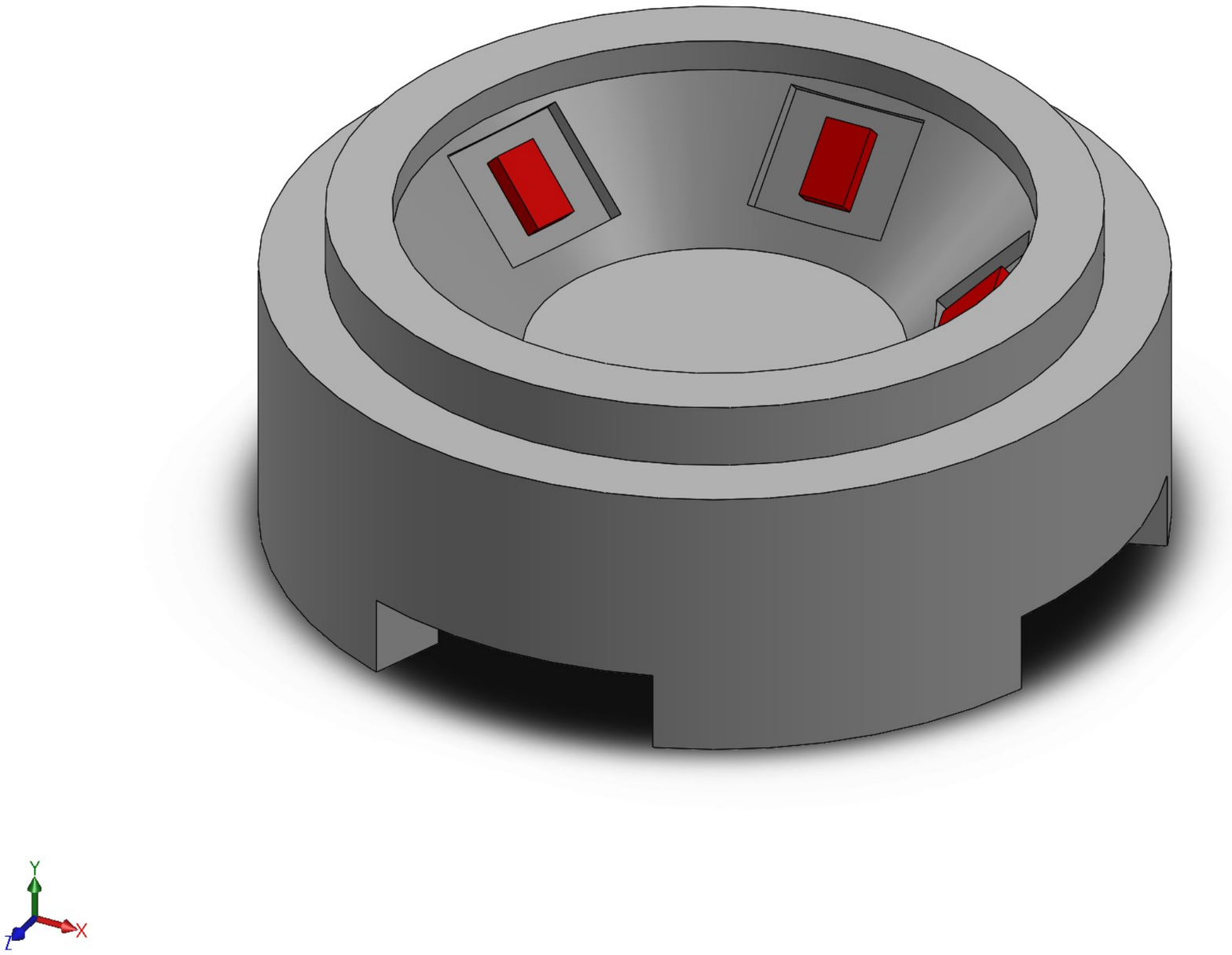}\\[0mm]
\footnotesize Case 6: Ring 25$^\circ$ & 
\footnotesize Case 7: Ring 30$^\circ$ & 
\footnotesize Case 8: Ring 35$^\circ$ &
\footnotesize Case 9: Ring 40$^\circ$ & 
\footnotesize Case 10: Ring 45$^\circ$ \\[0mm]
\end{tabular}
\caption{Trimetric view of all dome design cases. Takktile sensors are shown in red color and the volume of soft material is not represented.}
\label{fig:cases_cads}
\end{figure*}

\section{Exploration of sensor configurations using simulation}

\begin{figure*}[t]
\setlength{\tabcolsep}{1mm}
\begin{tabular}{ccccc}
\includegraphics[clip, trim=0.4cm 2.2cm 0.5cm 2.8cm, width=0.19\textwidth]{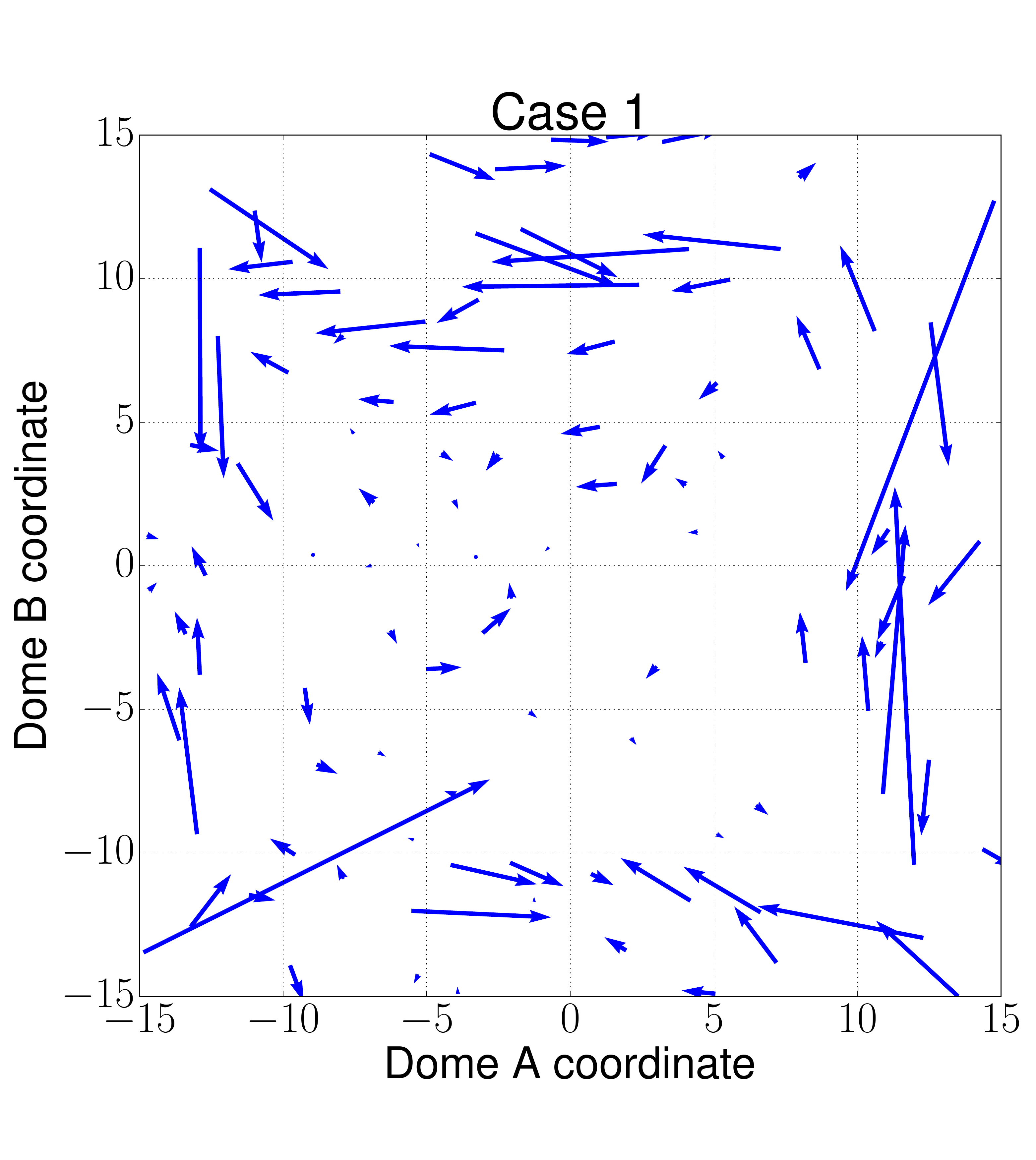}&
\includegraphics[clip, trim=0.4cm 2.2cm 0.5cm 2.8cm, width=0.19\textwidth]{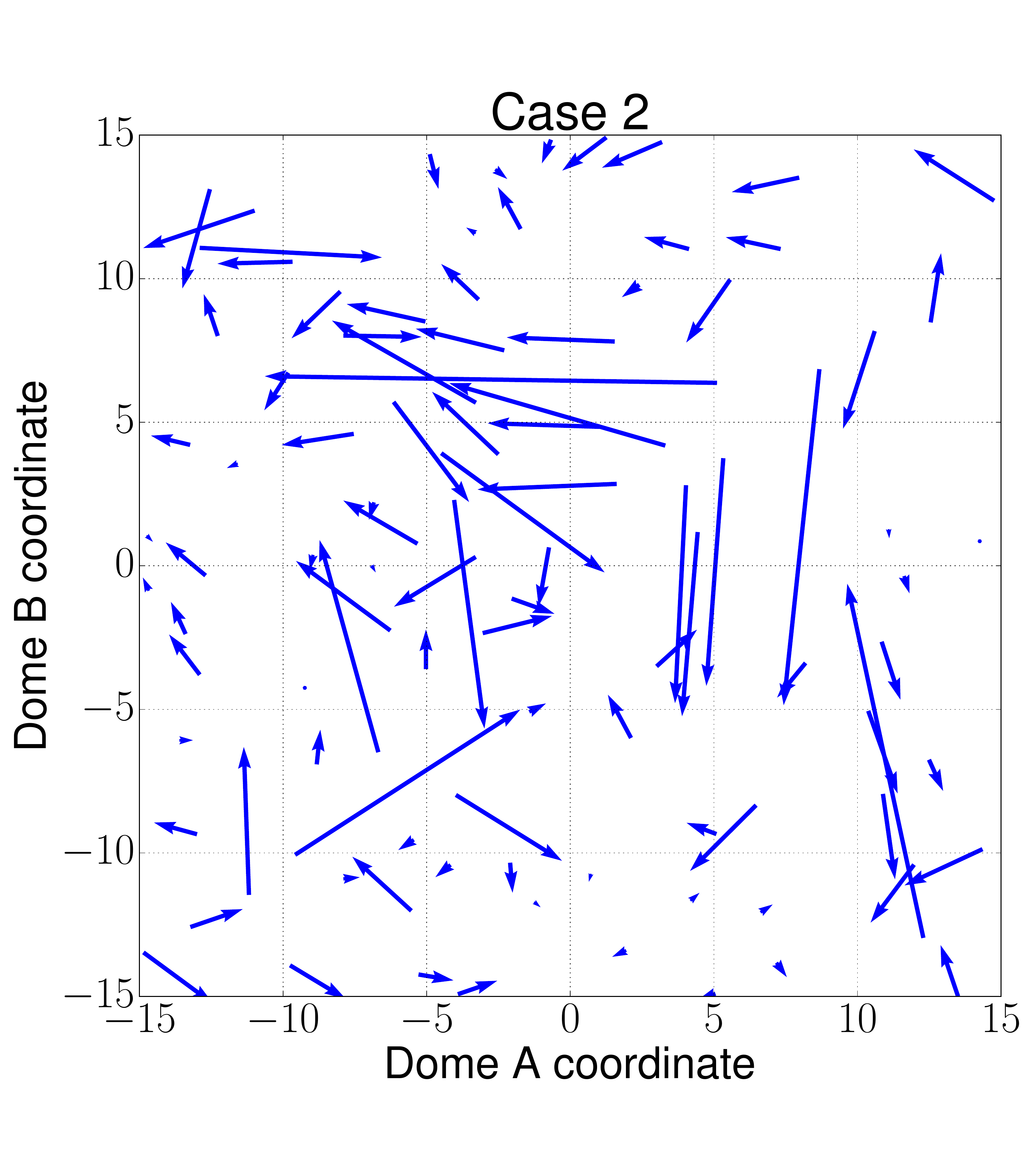}&
\includegraphics[clip, trim=0.4cm 2.2cm 0.5cm 2.8cm, width=0.19\textwidth]{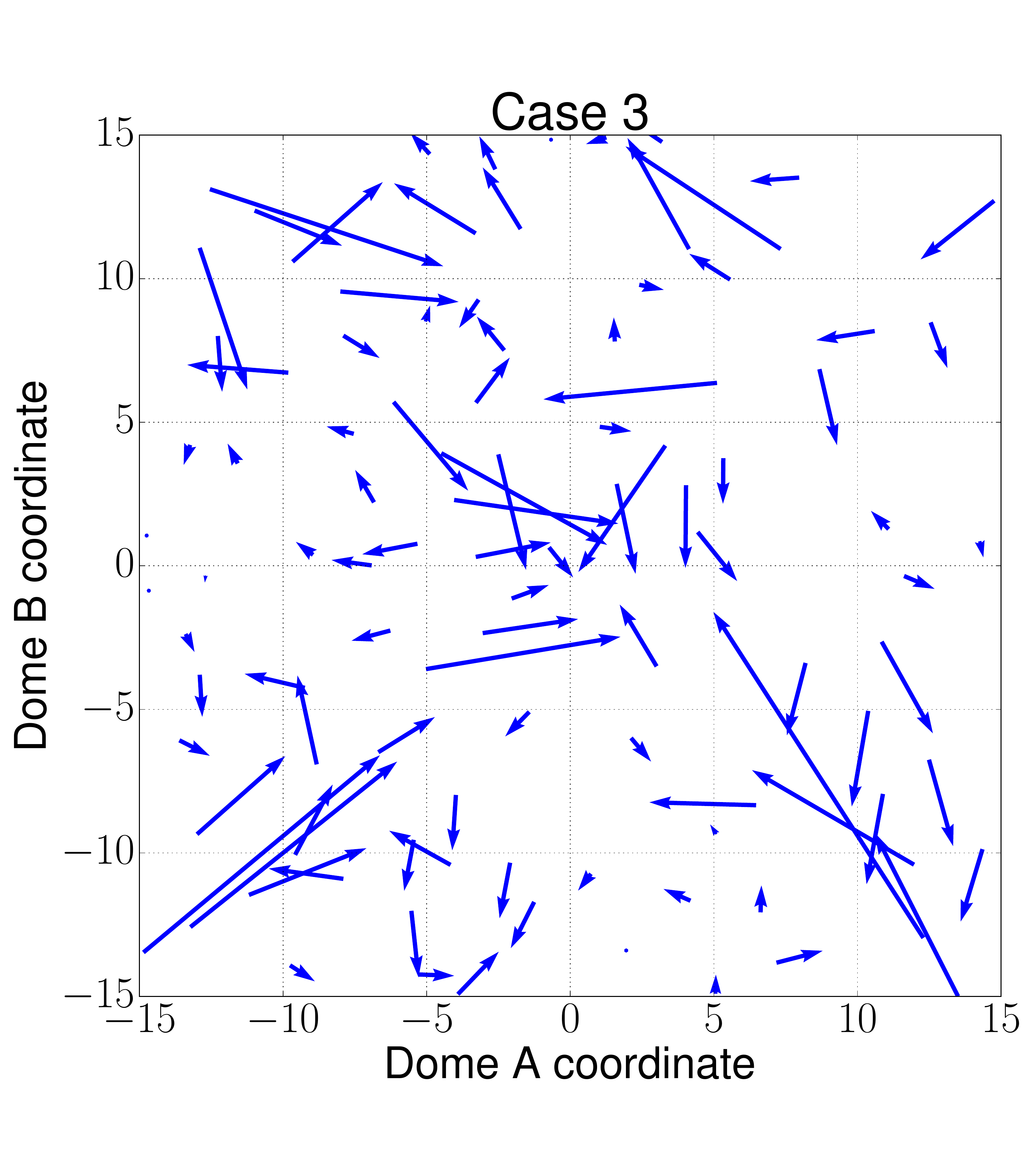}&
\includegraphics[clip, trim=0.4cm 2.2cm 0.5cm 2.8cm, width=0.19\textwidth]{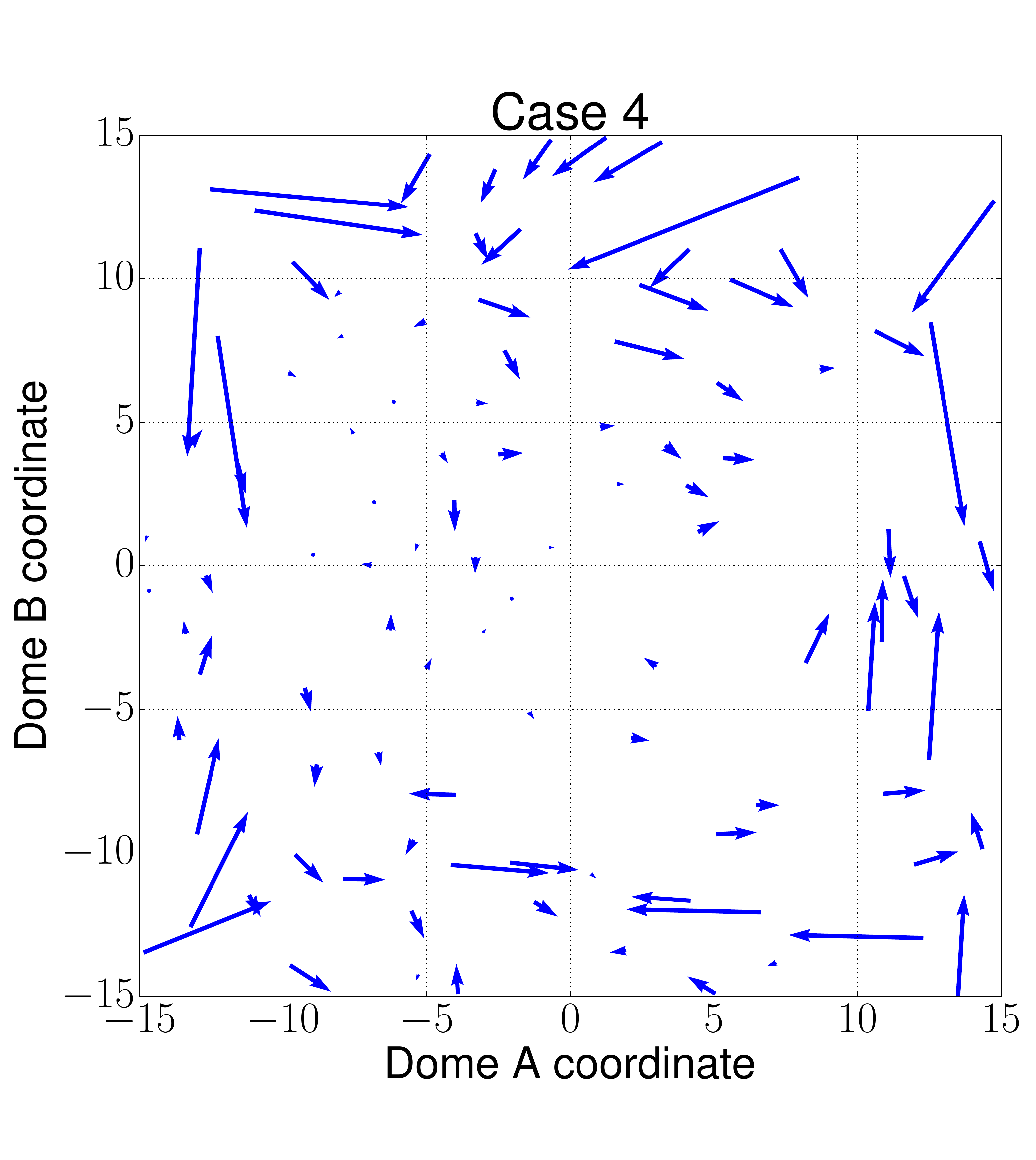}&
\includegraphics[clip, trim=0.4cm 2.2cm 0.5cm 2.8cm, width=0.19\textwidth]{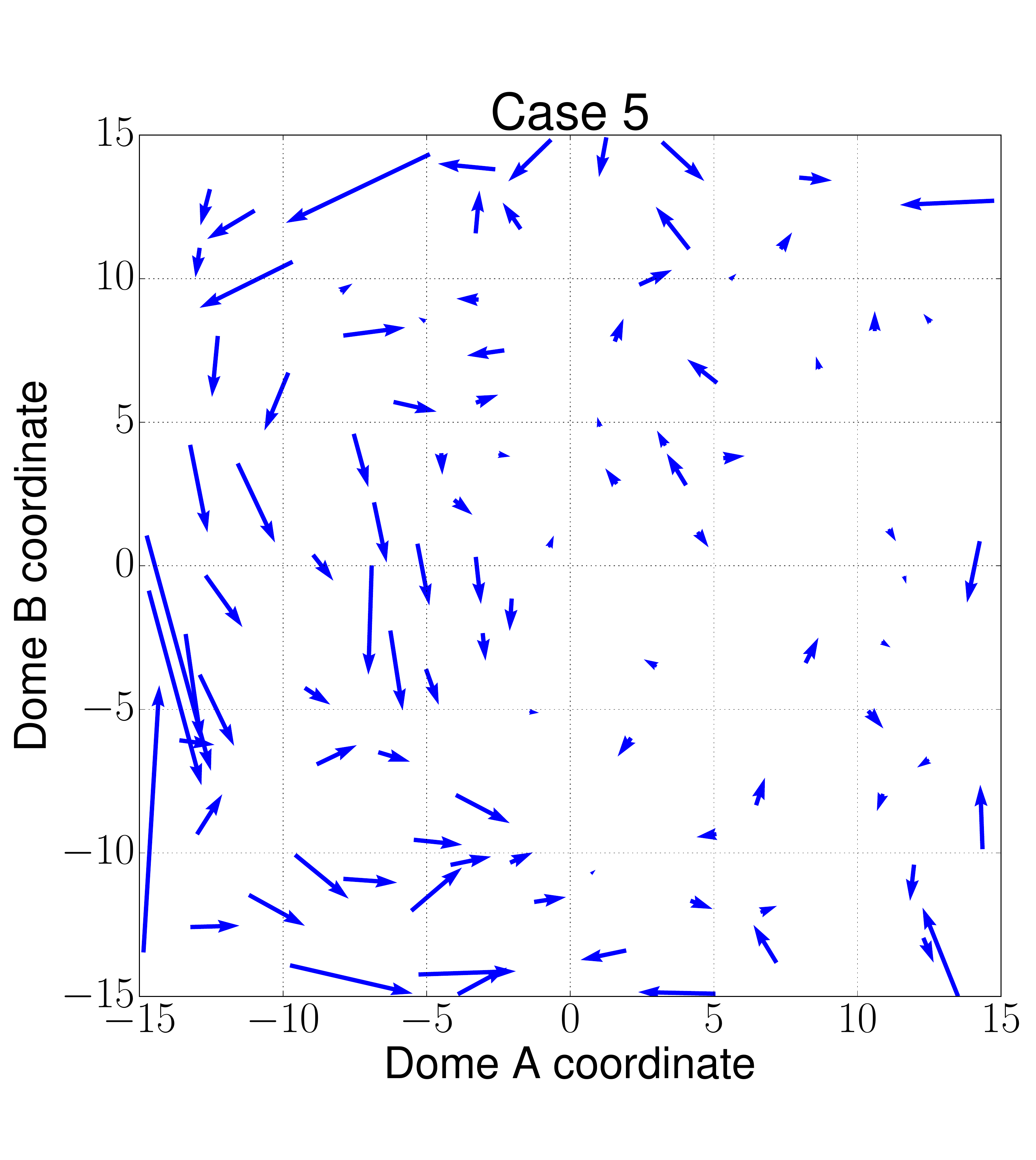}\\[1mm]
\includegraphics[clip, trim=0.4cm 2.2cm 0.5cm 2.8cm, width=0.19\textwidth]{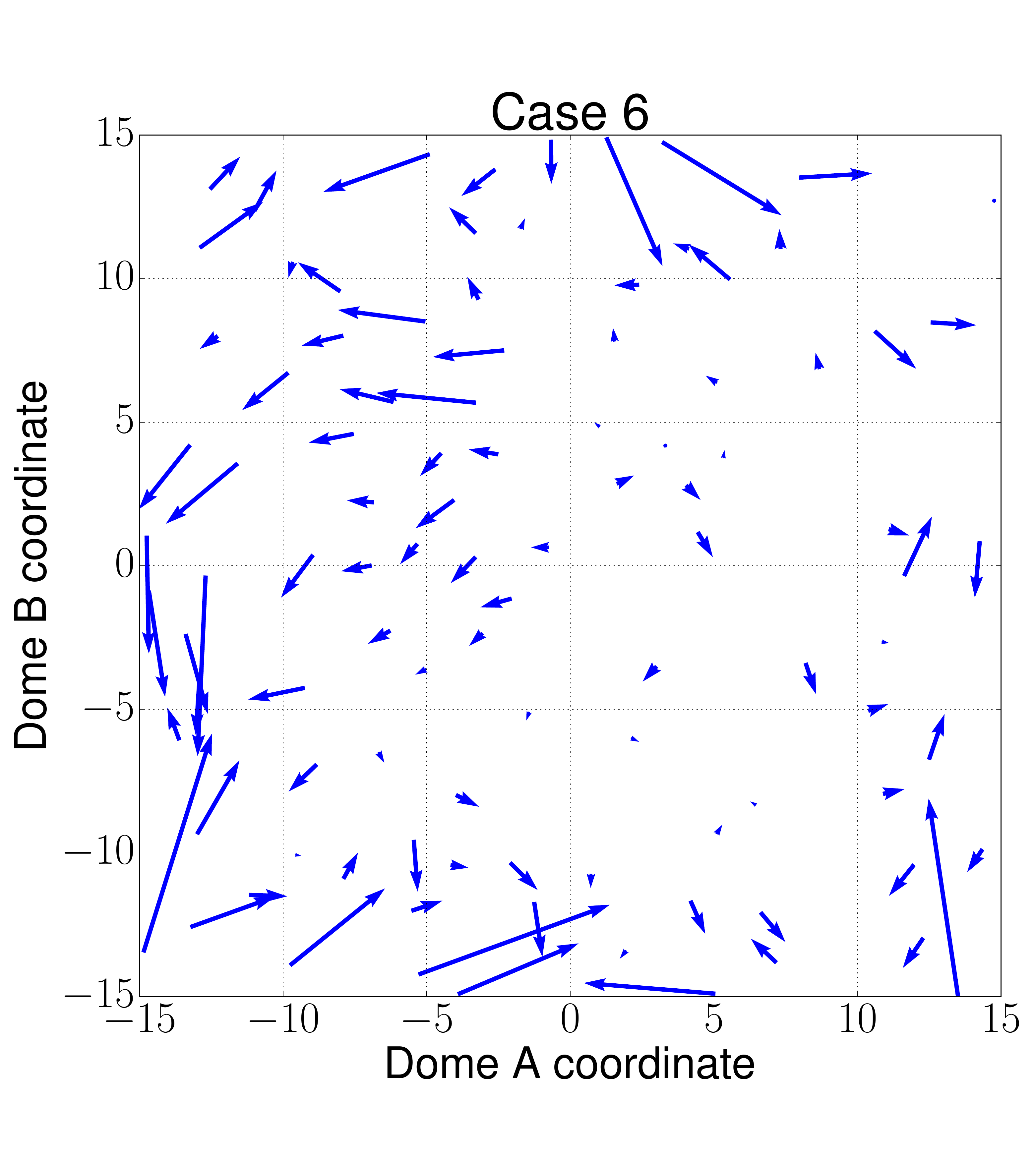}&
\includegraphics[clip, trim=0.4cm 2.2cm 0.5cm 2.8cm, width=0.19\textwidth]{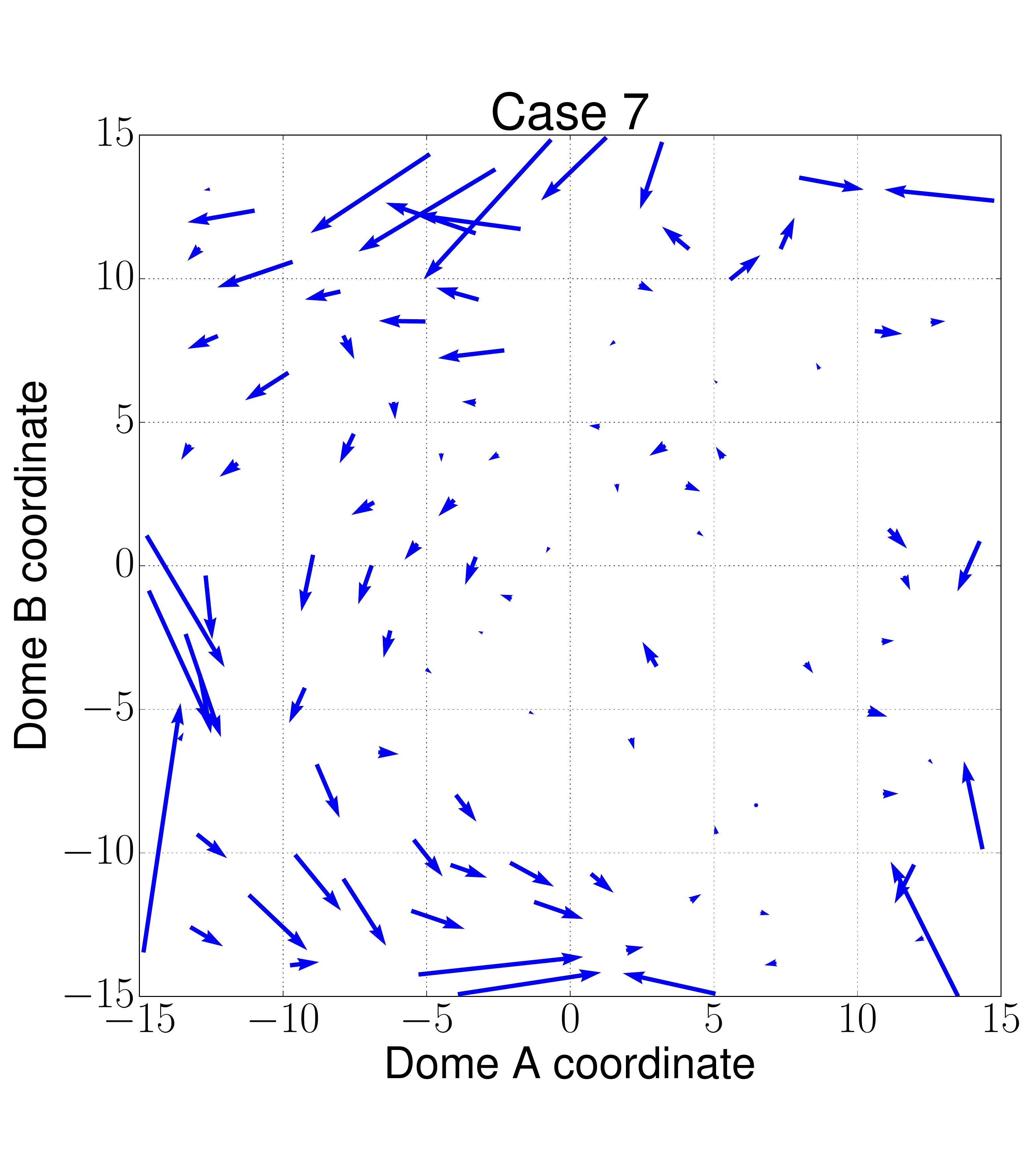}&
\includegraphics[clip, trim=0.4cm 2.2cm 0.5cm 2.8cm, width=0.19\textwidth]{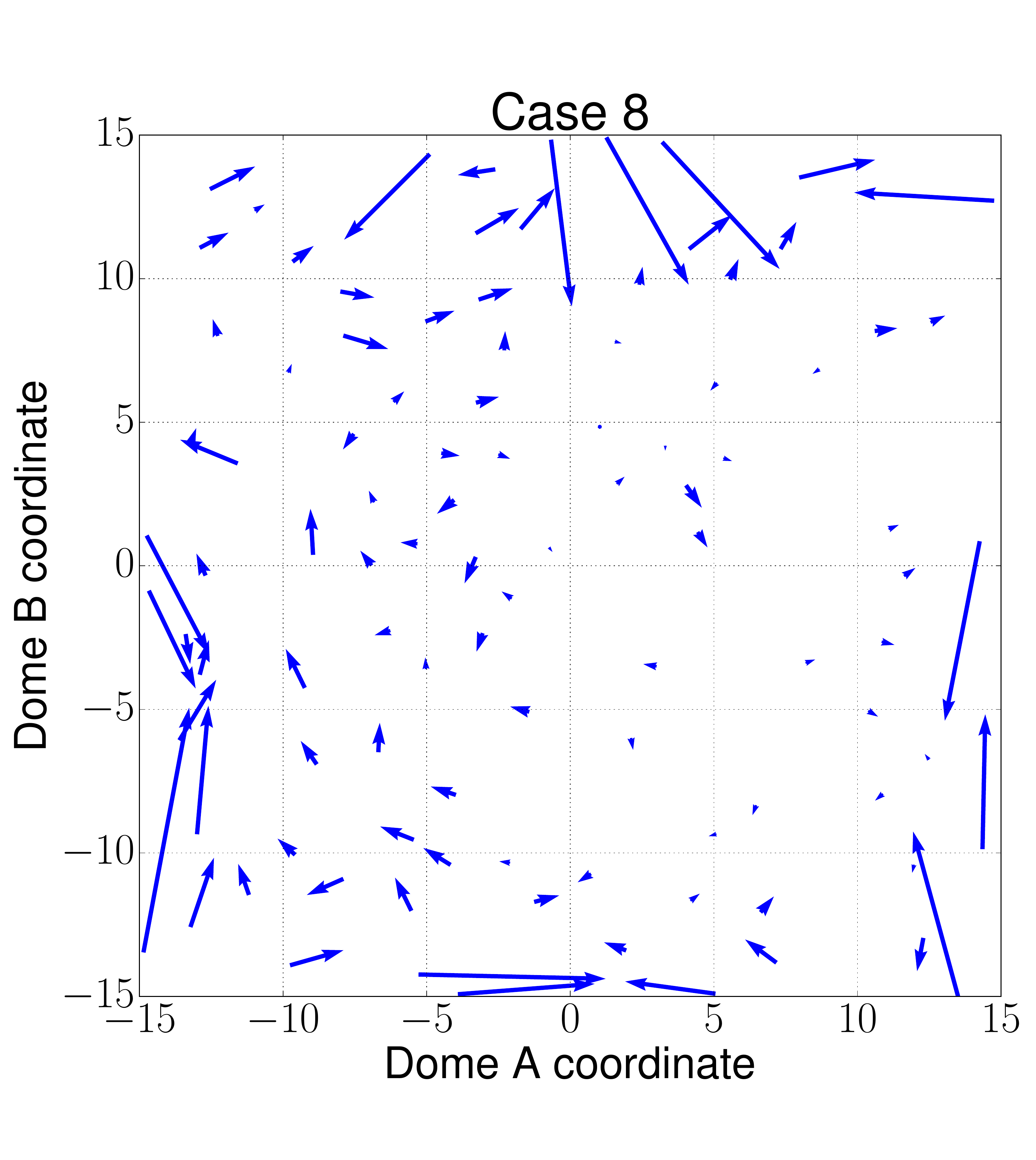}&
\includegraphics[clip, trim=0.4cm 2.2cm 0.5cm 2.8cm, width=0.19\textwidth]{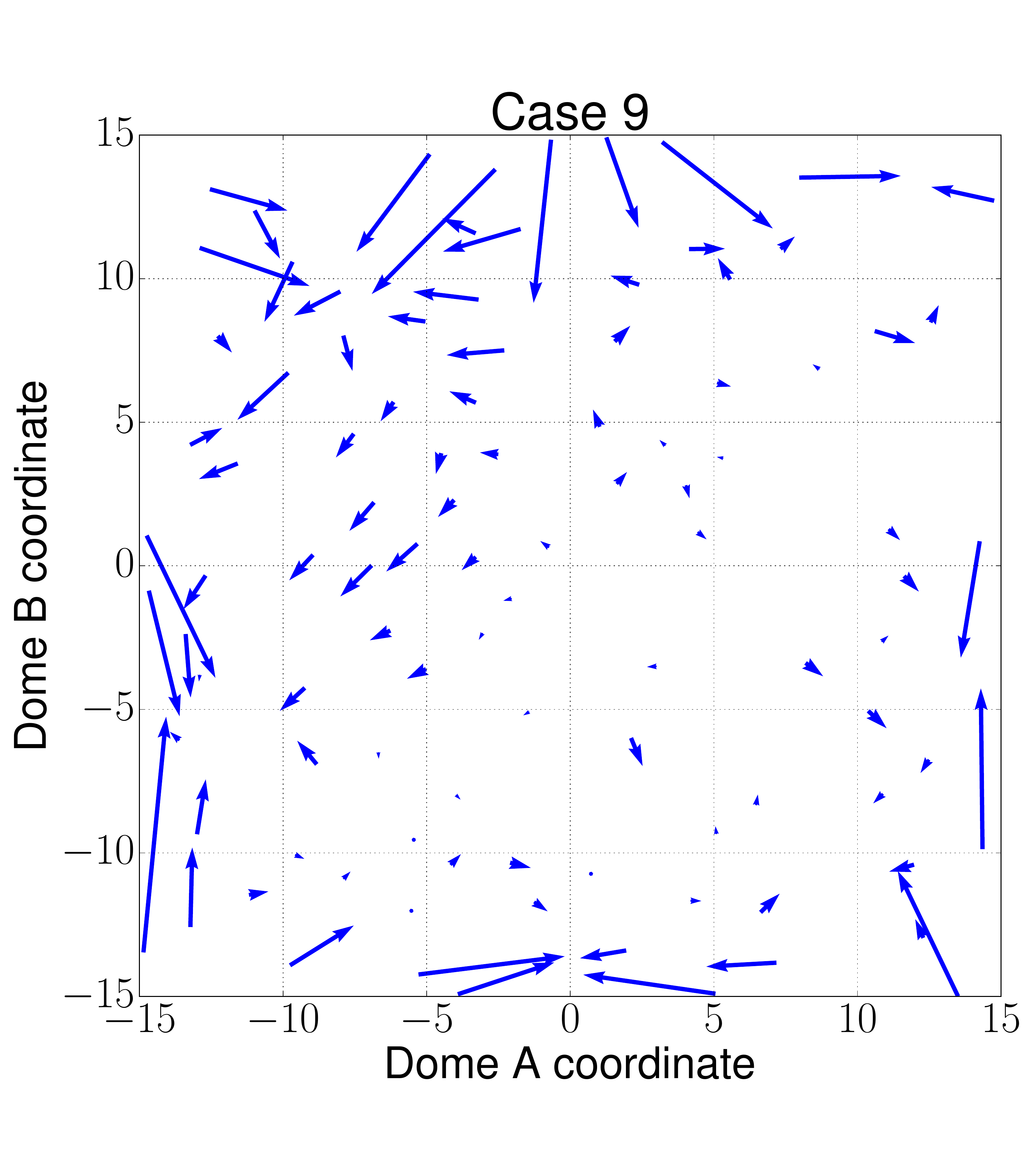}&
\includegraphics[clip, trim=0.4cm 2.2cm 0.5cm 2.8cm, width=0.19\textwidth]{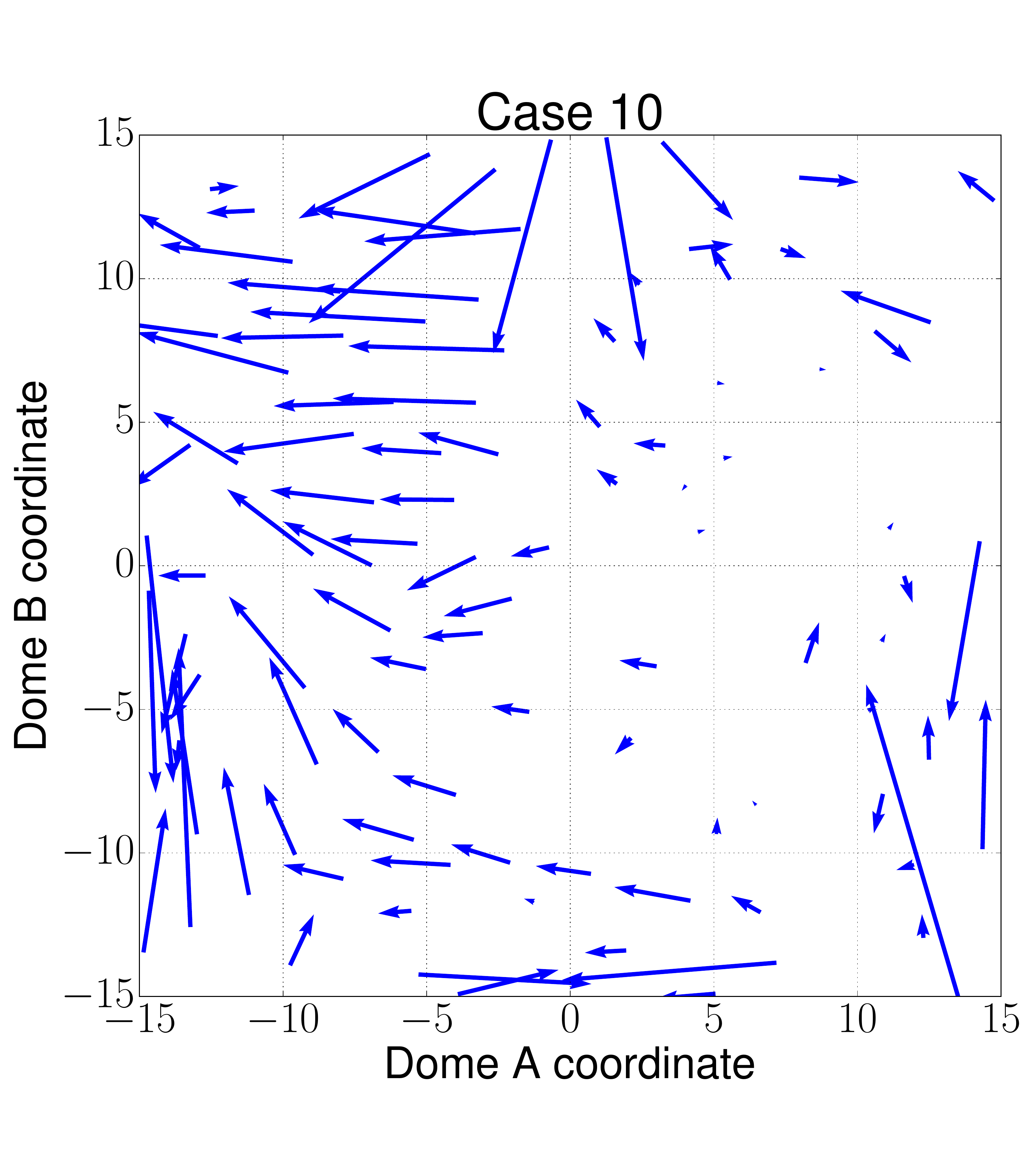}
\end{tabular}
\caption{Simulation localization results for all ten defined cases of Takktile sensor configurations. Each arrow represents one indentation in our test set; the base is at the ground truth location while the tip of the arrow shows the predicted location.}
\label{fig:localization}
\end{figure*}

Our data-driven approach allows complete freedom in choosing sensor
configurations inside the tactile dome and the simulation framework
allows us to test and compare different configurations without
building physical sensors. The space of possible configurations
is so large that the process of selecting configurations to test must
still be guided by intuition. In this study, we test 10 possible configurations, shown in
Figure~\ref{fig:cases_cads}. The performance of each case is summarized in Table~\ref{table1}, showing median, mean
and standard deviation for localization error in cartesian coordinates over the 100 test
indentations. Localization error plots for
each configuration are also presented in Figure~\ref{fig:localization}
in $(A,B)$ space for a more convenient 2D representation.

\begin{figure}[t]
\setlength{\tabcolsep}{0mm}
\begin{tabular}{cc}
\subfloat[Base with platform] {\label{case1}
\includegraphics[clip, trim=46cm 78cm 6cm 24cm, width=0.48\columnwidth]{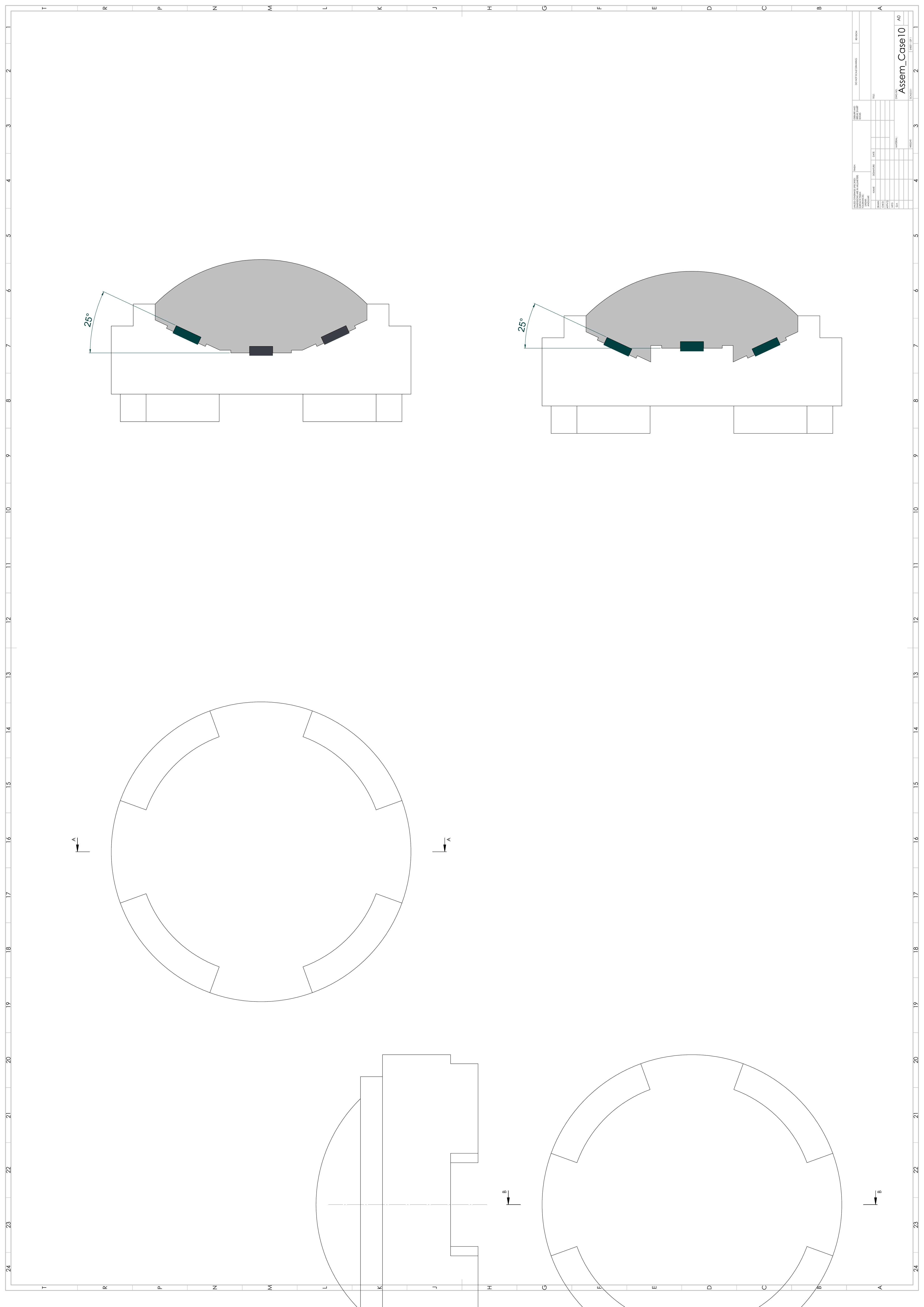}}\hfill &  \subfloat[Base without platform] {\label{case4}
\includegraphics[clip, trim=6cm 79cm 45cm 23cm, width=0.49\columnwidth]{images/Dome_section.pdf}}\\
\end{tabular}
\caption{Cross section of the dome illustrating a base (white) with platform (\ref{case1}, corresponds to Case 1) and one without platform (\ref{case4}, corresponds to Case 4). The mounting angle for both cases is positive 25 degrees. Takktile sensors are shown in black and the soft material is shown in gray.}
\label{fig:dome_slice}
\end{figure}

Our initial configuration (Case 1) is based on the intuition that we
want to promote sensor cross-talk: we want any given indentation to
result in some measurable signal in as many Takktile sensors as
possible. We define the ``mounting angle'' of the sensors as the angle
of the sensors relative to the floor where the dome rests (see
Figure~\ref{fig:dome_slice}). The angle is considered positive if the
Takktile sensor normal points towards the dome center. By this
convention, we hypothesize that a configuration with positive mounting
angles (concave) will provide better cross-talk when compared to one
with negative mounting angles (convex). On a convex mounting scheme,
the strain caused by an indentation on the edge of the dome can hardly
propagate to produce a meaningful signal on Takktile sensors on the
other side of the dome.

\begin{table}[t]
\centering
\caption{Localization accuracy results}
\label{table1}
\begin{tabular}{lccc}
\hline
\\[-3mm]
\multicolumn{1}{l}{\textbf{Case}} & \textbf{Median Err.} & \textbf{Mean Err.} & \textbf{Std. Dev.} \\ \hline
\\[-3mm] \hline
\\[-2mm]
Case 1 simulation & 1.5 mm & 2.0 mm & 2.2 mm   \\ 
Case 2 simulation & 1.9 mm & 2.6 mm & 2.4 mm   \\
Case 3 simulation &  2.1 mm & 2.5 mm & 1.9 mm   \\
Case 4 simulation & 1.2 mm & 1.6 mm & 1.5 mm   \\
Case 5 simulation & 1.4 mm & 1.7 mm & 1.4 mm   \\
Case 6 simulation & 1.2 mm & 1.7 mm & 1.4 mm   \\
Case 7 simulation & 1.0 mm & 1.6 mm & 1.6 mm   \\
Case 8 simulation & 0.9 mm & 1.5 mm & 1.7 mm   \\
Case 9 simulation & 1.0 mm & 1.6 mm & 1.6 mm   \\
Case 10 simulation & 2.6 mm & 3.0 mm & 2.3 mm   \\
Case 1 real & 1.4 mm & 1.6 mm & 1.2 mm   \\
Case 8 real & 1.1 mm & 1.7 mm & 1.9 mm   \\
\end{tabular}
\end{table}

We test this hypothesis in Cases 1 through 3. These three cases use the same ``central'' configuration, but with mounting angles of -25, 0 and 25 degrees. For Case 1, we raise the central sensor such that its distance to the dome's surface is reduced. This is to adjust for the Takktile sensors' decrease in sensitivity as they are embedded deeper in the soft volume~\cite{TENZER14}. As a reference, this distance for Case 1 is equivalent to that of Case 2 (for the central sensor). Our results confirm our initial hypothesis, with Case 1 showing the highest accuracy at $1.5mm$ median error. Moreover, Case 3 displays the worse performance, also in line with our hypothesis. Localization plots for Case 3 show small error only when indenting directly over the Takktile sensors' locations. Case 1 presents overall better predictions, albeit with larger errors for indentations close to the dome's edges.

With our cross-talk hypothesis confirmed, we test two additional variations of Case 1 while maintaining the peripheral Takktile sensors' mounting angle at 25 degrees. Case 4 removes the central sensor platform and Case 6 explores a new ``ring'' configuration. As our results show, both of these changes yield practically the same improvement with a median error of $1.2mm$.

Cases 5 and 7 through 10 are used to optimize the mounting angle of the Takktile sensors. Altogether we test angles of 15, 25, 30, 35, 40 and 45 degrees for the ring configuration. Note that as we increase the mounting angle, the Takktile sensors' sampling hole distance to the dome's surface is also increased. In order to keep the Takktile sensors' normals pointing at the dome's surface and to maintain the distance to the dome's surface, the radial distance at which we mount these sensors increases with the mounting angle. Case 5, with an angle of 15 degrees, results in a decrease of performance when compared to Case 6 (25 degrees). Increasing the angle, however, does improve the overall performance, with Case 8 (35 degrees) showing the best localization accuracy at $0.9mm$ median error. Angles greater than 35 degrees present a drop in performance, as exhibited by Case 9 and 10. 

In general, the localization error plots in Figure~\ref{fig:localization} show that in our best performing cases, the predictions for indentations on the dome center are the most accurate whereas indentations at the very edge of the tactile dome display overall bigger prediction errors. Our hypothesis is that in these particular locations, the edges of the ABS base resolve the indenter force dominating the stress distribution and not enough sensors are excited to provide a better prediction.

\section{Physical sensor testing and validation}

In order to validate our simulation results, we build two physical tactile domes representing Case 1 and Case 8. We use two methods to compare the real sensor with the simulation. First, we look at the individual Takktile sensors response when we indent the dome along the symmetry lines defined in Figure~\ref{fig:distributions}. Second, we feed the real data to our localization algorithm and compare the localization error plots to evaluate their correlation both in terms of trends and numerical results. 

\begin{figure}[t]
\setlength{\tabcolsep}{1mm}
\begin{tabular}{cc}
\includegraphics[clip, trim=0.4cm 0.5cm 0.5cm 0.3cm, width=0.49\columnwidth]{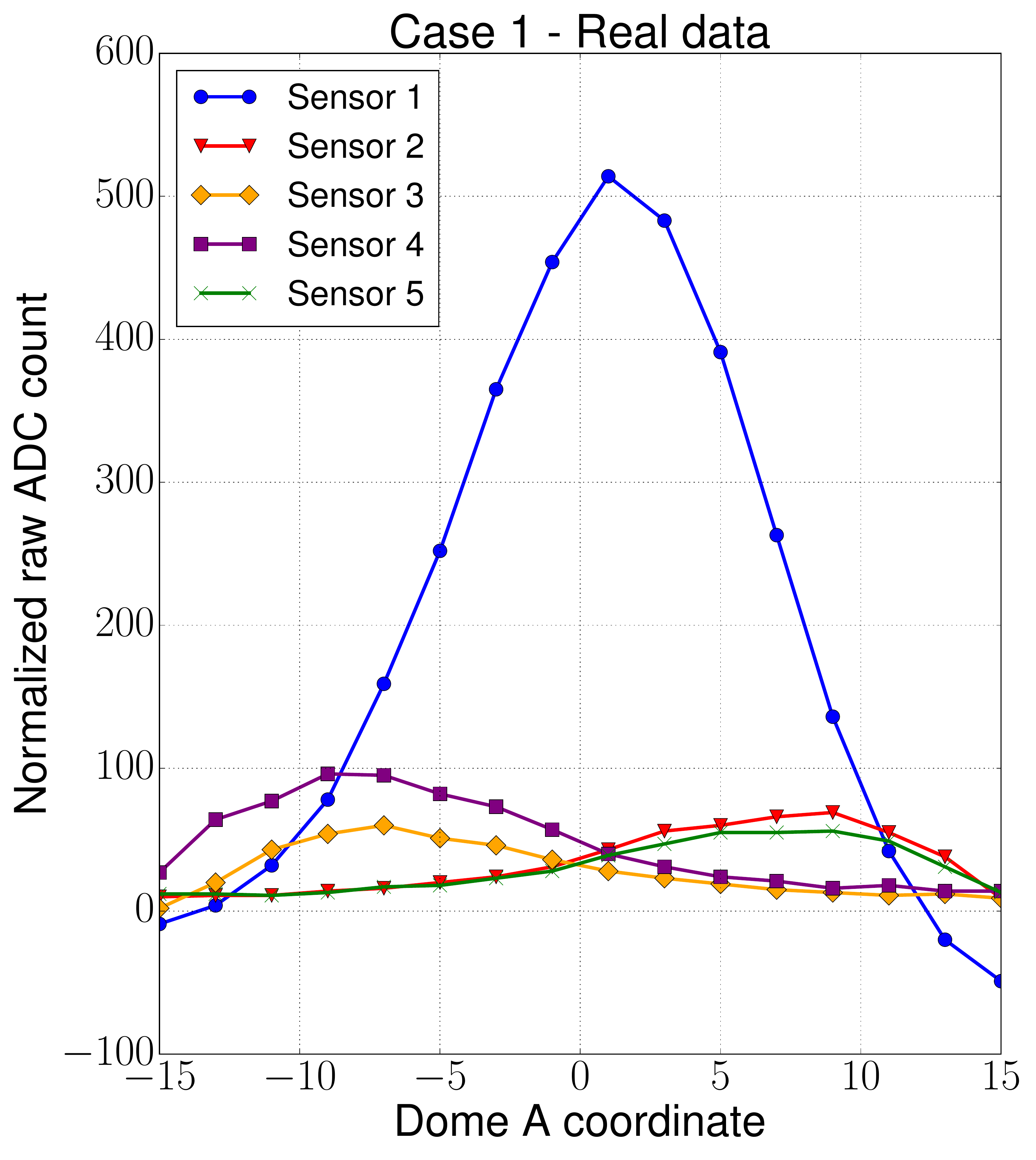}&
\includegraphics[clip, trim=0.4cm 0.5cm 0.5cm 0.3cm, width=0.49\columnwidth]{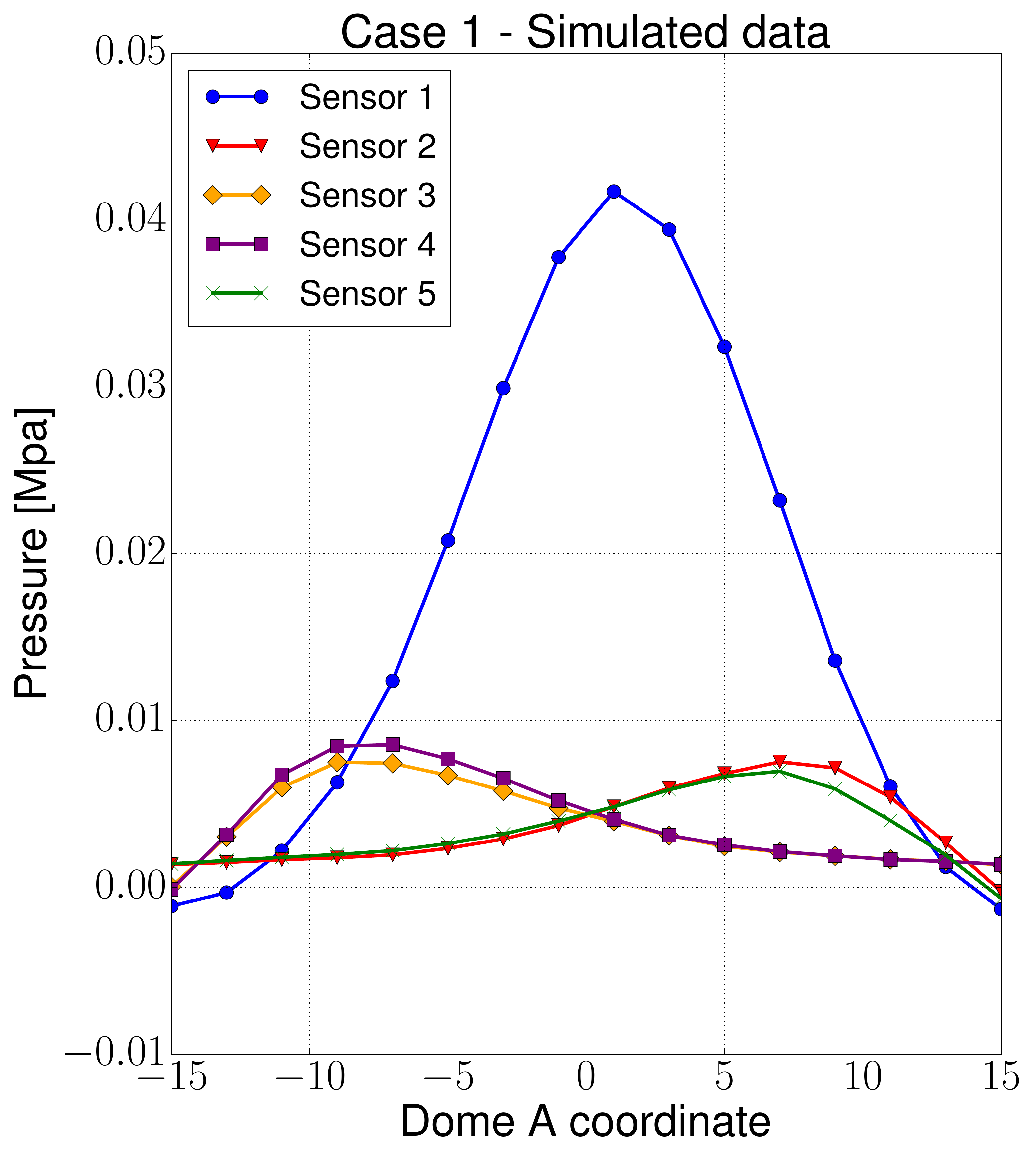}\\[1mm]
\includegraphics[clip, trim=0.4cm 2.2cm 0.5cm 2.8cm, width=0.49\columnwidth]{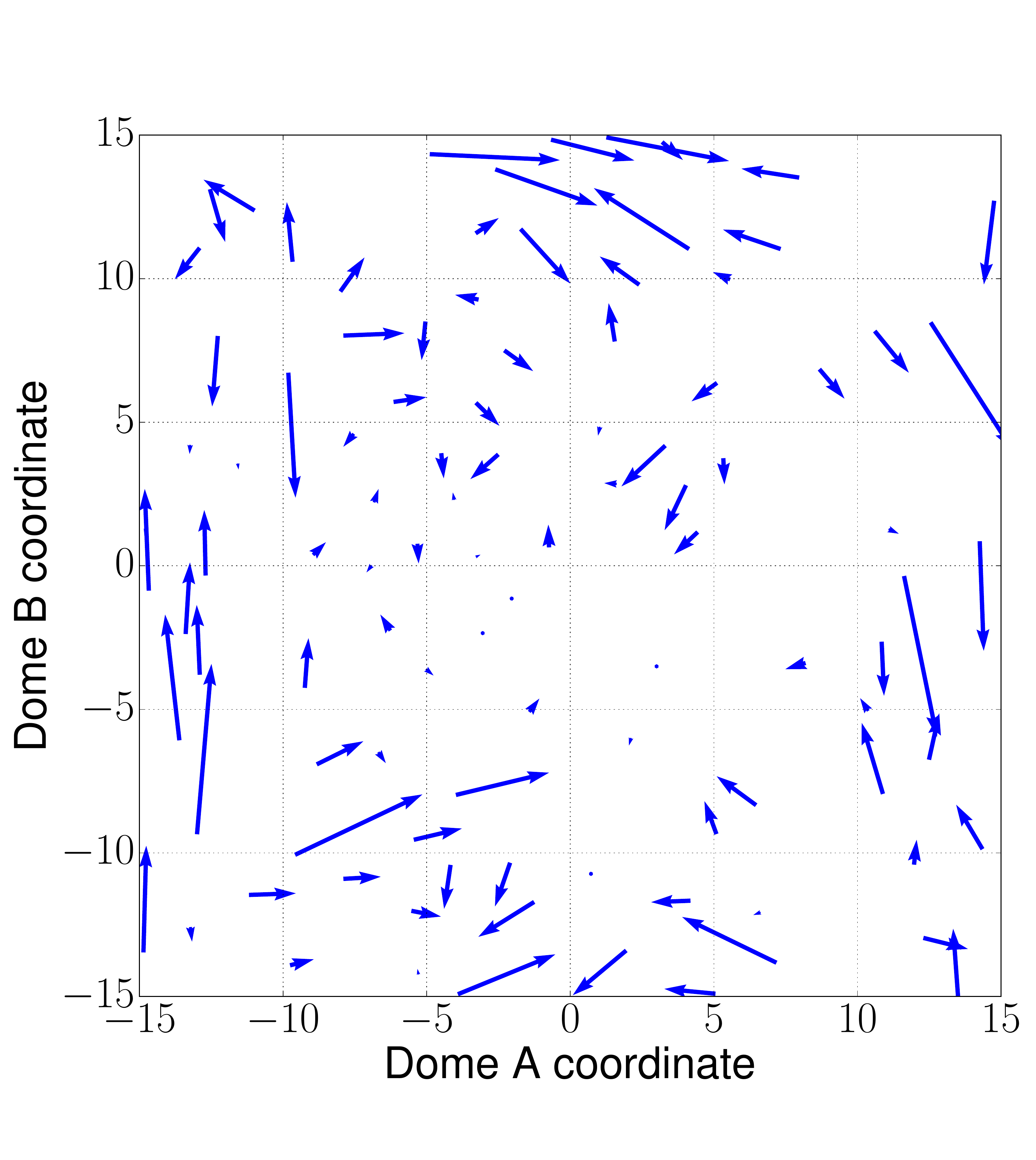}&
\includegraphics[clip, trim=0.4cm 2.2cm 0.5cm 2.8cm, width=0.49\columnwidth]{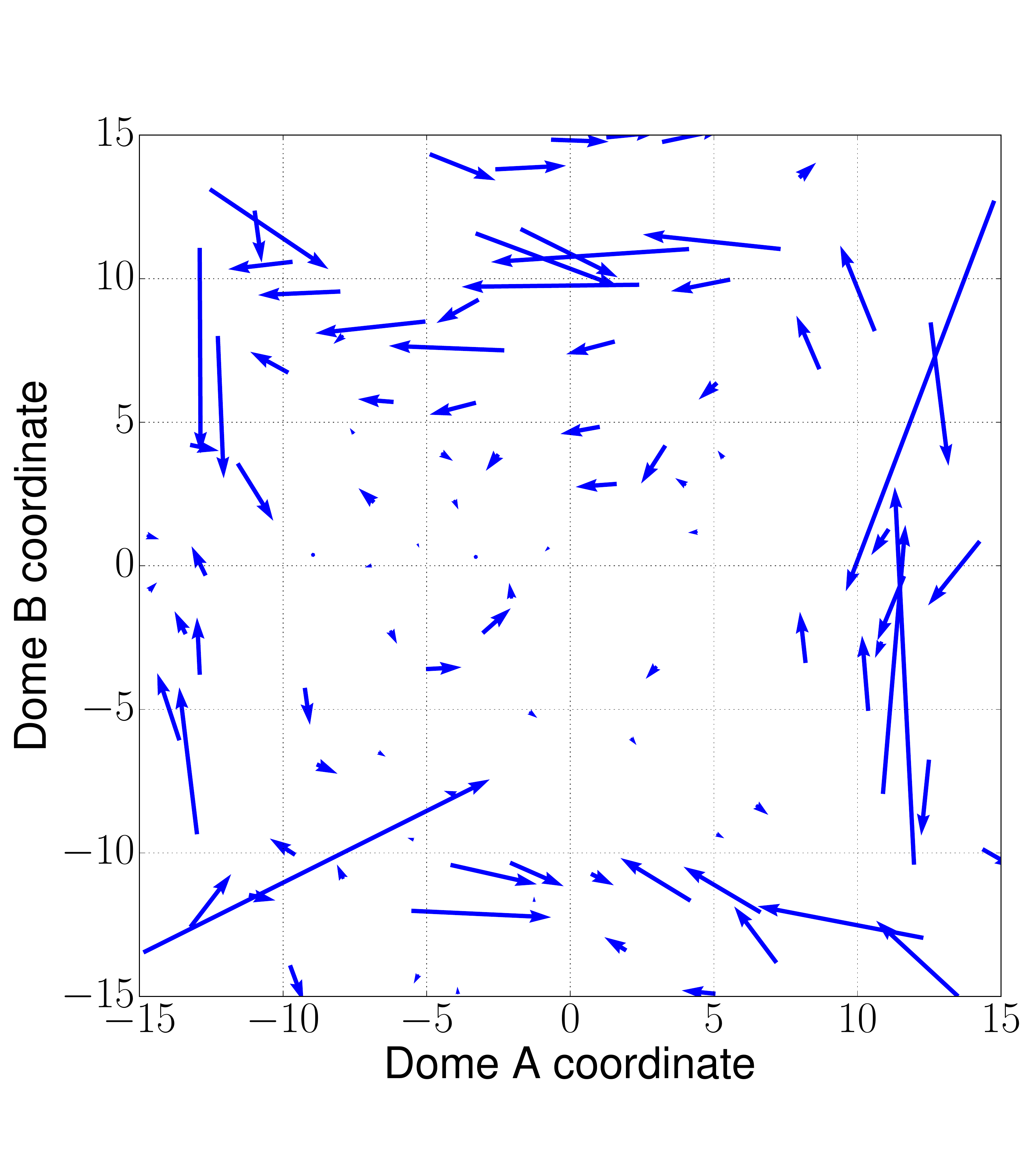}

\end{tabular}
\caption{Takktile sensors response when the dome is indented along symmetry line and localization plots comparison for Case 1. Left shows the data from the real sensor, right shows the simulation data.}
\label{fig:Case1_physical}
\end{figure}

Figure~\ref{fig:Case1_physical} shows these two evaluation methods for Case 1. The sensor response along the symmetry line displays the same trend for both the real data and the simulated data. Takktile sensor number two in the real data shows some asymmetry with respect to Takktile sensor four's response. This could be attributed to an incomplete degassing of this particular Takktile sensor.  For the localization error, both plots display a similar trend where large errors occur close to the dome edges. Numerically, the median error for the real case at $1.4mm$ is slightly better than what our simulation predicted ($1.5mm$). 

\begin{figure}[t]
\setlength{\tabcolsep}{1mm}
\begin{tabular}{cc}
\includegraphics[clip, trim=0.4cm 0.5cm 0.5cm 0.3cm, width=0.49\columnwidth]{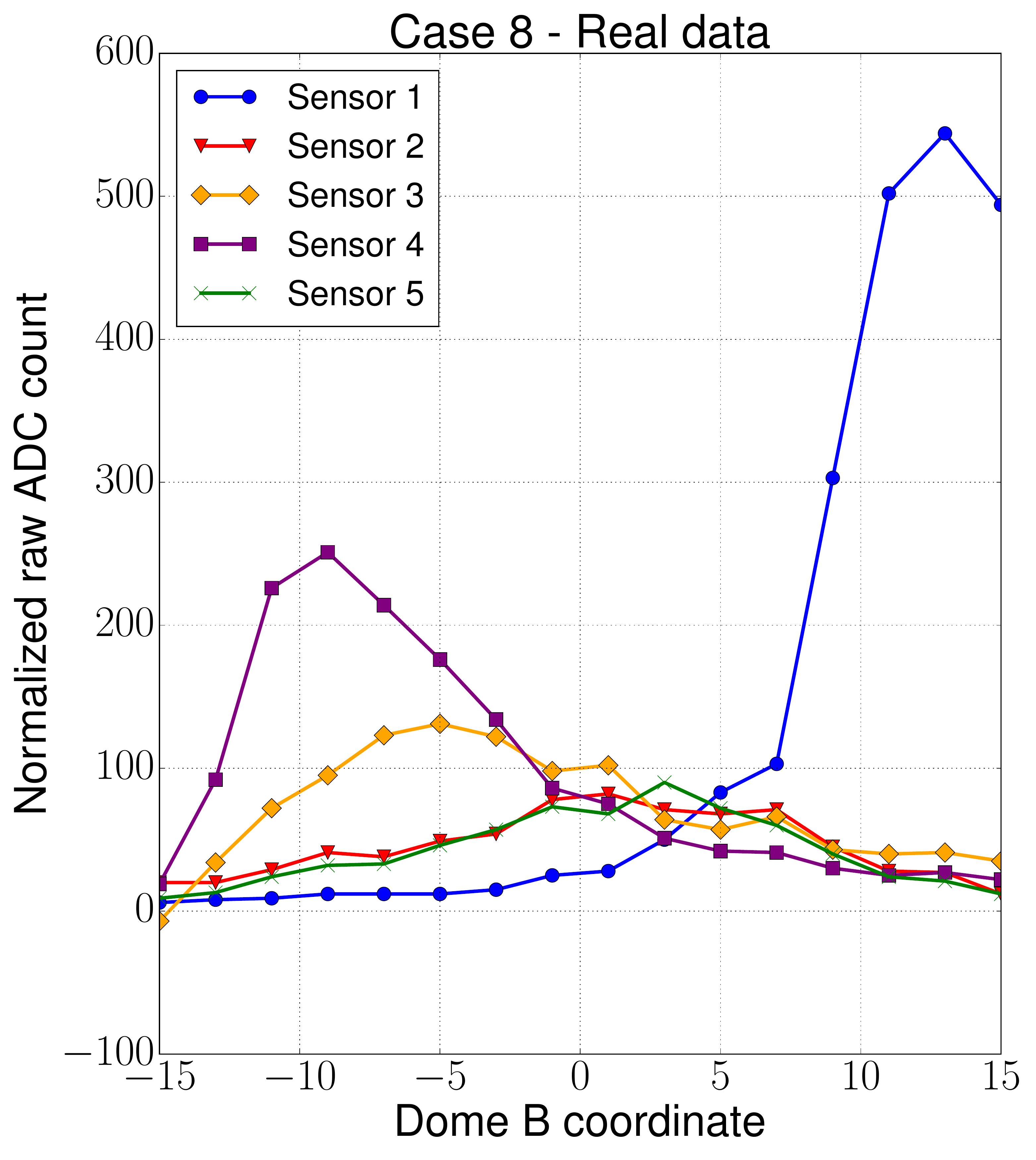}&
\includegraphics[clip, trim=0.4cm 0.5cm 0.5cm 0.3cm, width=0.49\columnwidth]{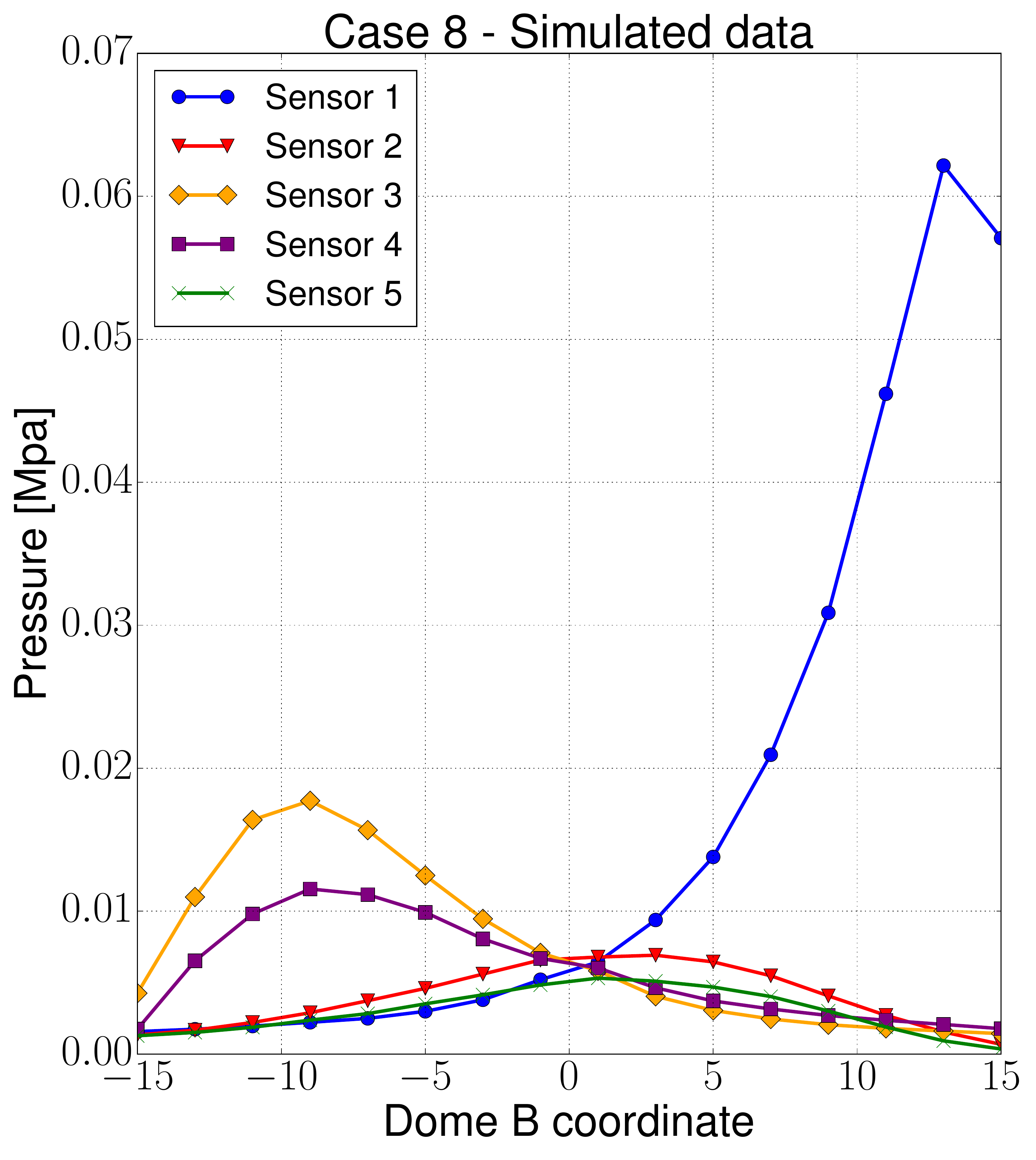}\\[1mm]
\includegraphics[clip, trim=0.4cm 2.2cm 0.5cm 2.8cm, width=0.49\columnwidth]{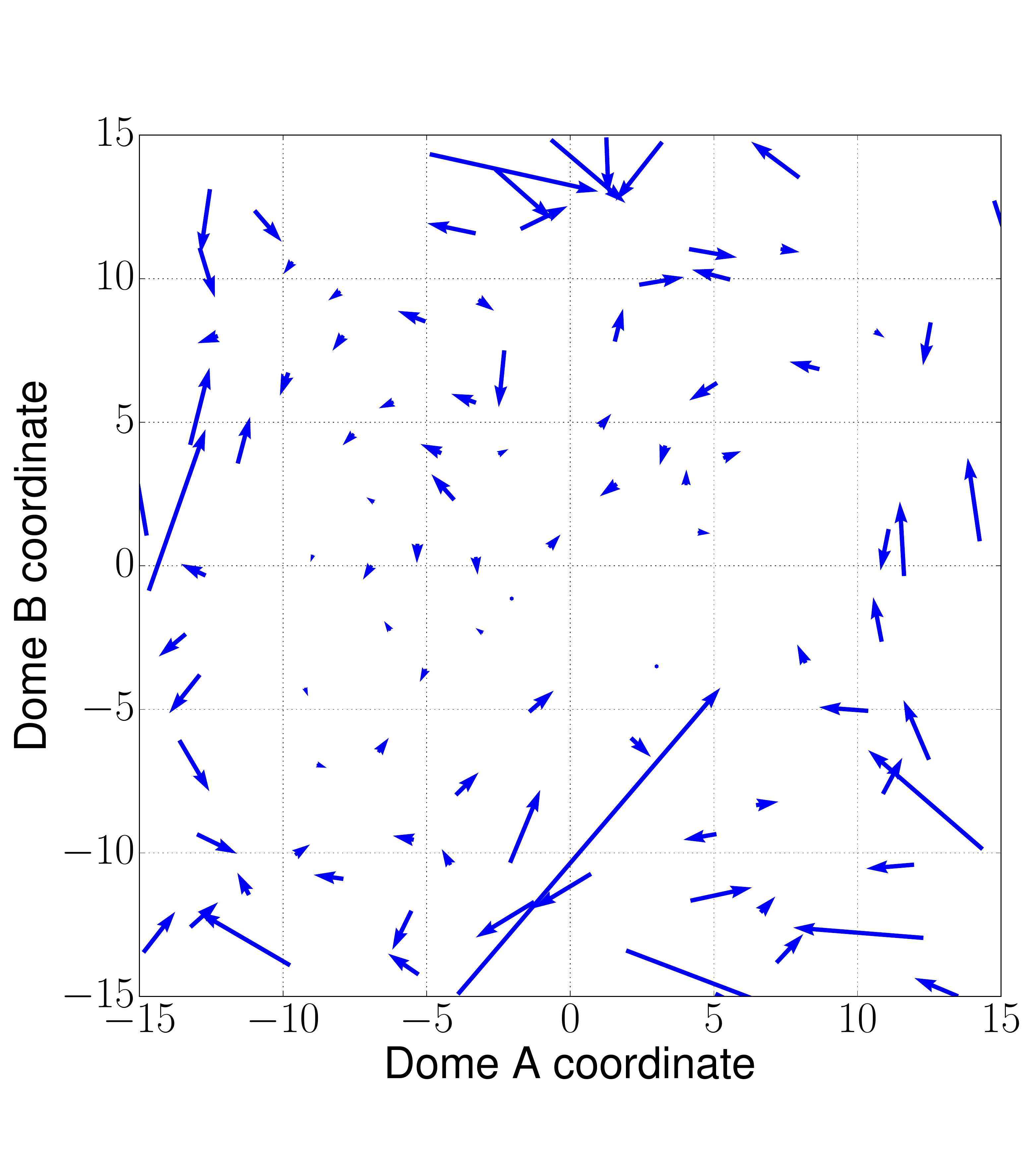}&
\includegraphics[clip, trim=0.4cm 2.2cm 0.5cm 2.8cm, width=0.49\columnwidth]{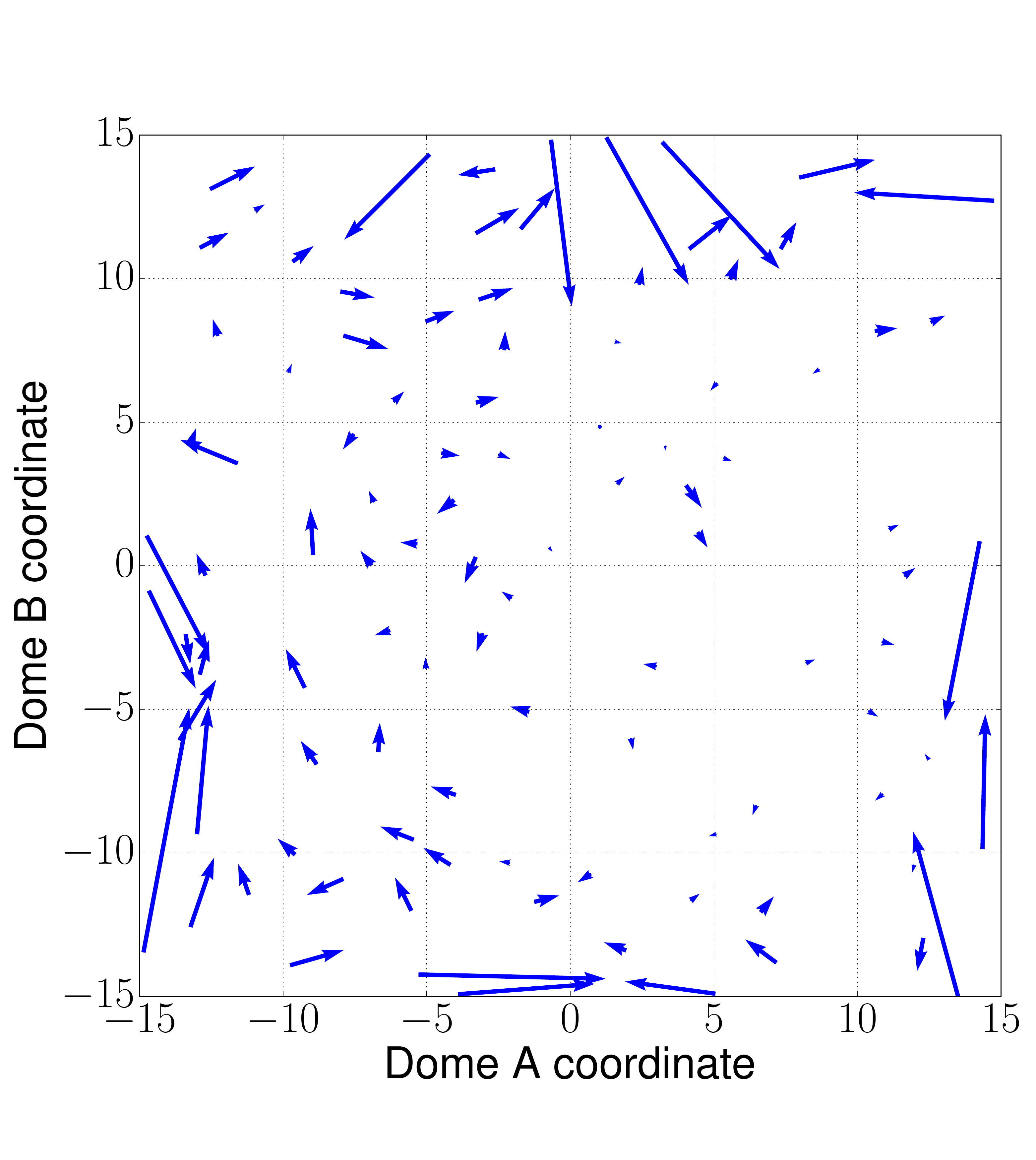}

\end{tabular}
\caption{Takktile sensors response when the dome is indented along symmetry line and localization plots comparison for Case 8. Left shows the data from the real sensor, right shows the simulation data.}
\label{fig:Case8_physical}
\end{figure}

For Case 8, the simulation shows a similar trend to the real data  when indenting along the symmetry line, although it does not correspond as well as Case 1 (see Figure~\ref{fig:Case8_physical}). It must be noted that this data represents an indentation along the axis $A=1$, slightly offset from the symmetry line ($A=0$), which we use to mirror the data obtained in simulation. While our real data performs slightly worse than our simulation predicted numerically, the sensor does outperform Case 1 with a median error of $1.1mm$. The real and simulated localization plots display the same pattern in that large errors occur close to the edges of the dome.

\section{Conclusion}

In this paper we present an approach for accurately localizing touch over curved, three dimensional surfaces. The core of our method is embedding sparse pressure sensors in a soft volume and using a purely data-driven approach to learn the mapping from these sensor readings to the location of the indentation. Our data-driven approach allows complete freedom in placing the sensors inside the volume. Our intuition is that, through cross-talk between sensors, each indentation location can produce a signature that will allow for precise localization. However, not all sensor configurations will yield the same performance; we thus propose the use of simulation to explore possible designs. 

We validate this approach by simulating and building a hemispherical tactile dome made of soft material with five pressure sensors embedded within its volume. Our results show that this tactile dome is capable of localizing touch over its three dimensional surface. We use a custom built indenter machine to collect data by indenting the tactile dome at 256 different locations along its surface. Subsequently, we mine this data to learn the mapping between individual pressure sensor signals and the location where the indentation occurred. Our tactile dome can localize touch with a median error of $1.1mm$ over a surface of approximately $1300mm^2$ using only five pressure sensors. This results in a low-cost, easy to manufacture sensor capable of providing tactile feedback over a three dimensional surface.

To achieve these results, we use FEA to test multiple sensor configurations and in the process, validate the results of the simulation against real data. Our simulation exhibits the same trends as the real sensors in both pressure distribution among the embedded pressure sensors and overall performance of the localization algorithm. The FEA simulation proves to be a valuable tool for exploring the design space for a tactile system of these characteristics. It enables us to test the pressure sensors placement inside the volume of material and to effectively predict the final performance of the system. However, it is worth highlighting that even with simulation, it is impossible to completely explore the design space. We still rely on intuition to select candidate designs. The placement of sensors can have counter intuitive effects on the final performance; for example, having a raised central sensor in Case 1 does not provide better localization on the central region when compared to other cases.

Our final design for the tactile dome, based on Case 8, exhibits $1.1mm$ median error in localization accuracy with most large errors located at the very edge of the dome. It is likely that the base edges affect the stress distribution of the indentation. While not shown in this initial prototype, this trait might be avoided by a design with the Takktile sensors holding structure acting as a skeleton, and rubber cast on top of it, thus avoiding hard edges altogether and providing full coverage. 

Our data-driven methodology presents some limitations. Our model is built with real world data subject to certain conditions. The resulting model is constructed for the specific indenter shape used, with indentations constrained to directions normal to the sensor dome at fixed indentation depths. Therefore, it is uncertain what our model would predict if we were to use a different indenter shape, indent at several locations at once (multitouch), or at an angle other than the surface normal. We expect that given appropriate training data, a data-driven algorithm will be able to learn to predict at least a subset of these parameters. Future work will test this hypothesis by leveraging our simulated model to deploy a larger number of sensors in an artificial finger with the goal of predicting additional touch parameters (e.g. normal and shear forces, or the direction of the indentation with respect to the surface normal). Other important factors not considered in this preliminary study include hysteresis, the effect of environmental factors, and frequency response characterization. 

One important design parameter not explored in this study is the nature of the material used to fabricate the tactile dome. In this study we use Vytaflex 20 since it is the material used by the manufacturer to cast Takktile sensors. An interesting direction of research is to study how different material properties, like hardness or Poisson's ratio,  affect the pressure distribution for a given design. This would open the door to specially engineered materials to manufacture tactile sensors. We intend to further explore this possibility in the future.

\bibliographystyle{IEEEtran}
\bibliography{bib/grasping,bib/tactile,bib/thesis}

\begin{thebibliography}{10}
\providecommand{\url}[1]{#1}
\csname url@rmstyle\endcsname
\providecommand{\newblock}{\relax}
\providecommand{\bibinfo}[2]{#2}
\providecommand\BIBentrySTDinterwordspacing{\spaceskip=0pt\relax}
\providecommand\BIBentryALTinterwordstretchfactor{4}
\providecommand\BIBentryALTinterwordspacing{\spaceskip=\fontdimen2\font plus
\BIBentryALTinterwordstretchfactor\fontdimen3\font minus
  \fontdimen4\font\relax}
\providecommand\BIBforeignlanguage[2]{{%
\expandafter\ifx\csname l@#1\endcsname\relax
\typeout{** WARNING: IEEEtran.bst: No hyphenation pattern has been}%
\typeout{** loaded for the language `#1'. Using the pattern for}%
\typeout{** the default language instead.}%
\else
\language=\csname l@#1\endcsname
\fi
#2}}

\bibitem{DAHIYA10}
R.~Dahiya, G.~Metta, M.~Valle, and G.~Sandini, ``Tactile sensing: From humans
  to humanoids,'' \emph{IEEE Trans. on Robotics}, vol.~26, no.~1, 2010.

\bibitem{HAMMOCK13}
M.~Hammock, A.~Chortos, B.~Tee, J.~Tok, and Z.~Bao, ``The evolution of
  electronic skin (e‐skin): A brief history, design considerations, and
  recent progress,'' \emph{Advanced Materials}, vol.~25, no.~42, pp.
  5997--6038, 2013.

\bibitem{kappassov2015}
Z.~Kappassov, J.-A. Corrales, and V.~Perdereau, ``Tactile sensing in dexterous
  robot hands — review,'' \emph{Robotics and Autonomous Systems}, vol.~74,
  pp. 195--220, 2015.

\bibitem{JENTOFT14}
L.~Jentoft, Q.~Wan, and R.~Howe, ``Limits to compliance and the role of tactile
  sensing in grasping,'' in \emph{IEEE Int. Conf. on Robotics and Automation},
  2014.

\bibitem{WAN16}
Q.~Wan, R.~P. Adams, and R.~Howe, ``Variability and predictability in tactile
  sensing during grasping,'' in \emph{IEEE Int. Conf. on Robotics and
  Automation}, 2016.

\bibitem{TENZER14}
Y.~Tenzer, L.~Jentoft, and R.~Howe, ``The feel of mems barometers: Inexpensive
  and easily customized tactile array sensors,'' \emph{IEEE Robotics \&
  Automation Magazine}, vol.~21, no.~3, 2014.

\bibitem{reeks2016angled}
C.~Reeks, M.~G. Carmichael, D.~Liu, and K.~J. Waldron, ``Angled sensor
  configuration capable of measuring tri-axial forces for phri,'' in
  \emph{Robotics and Automation (ICRA), 2016 IEEE Int. Conf. on}.\hskip 1em
  plus 0.5em minus 0.4em\relax IEEE, 2016.

\bibitem{guggenheim2017}
J.~W. Guggenheim, L.~P. Jentoft, Y.~Tenzer, and R.~D. Howe, ``Robust and
  inexpensive six-axis force--torque sensors using mems barometers,''
  \emph{IEEE/ASME Transactions on Mechatronics}, vol.~22, no.~2, pp. 838--844,
  2017.

\bibitem{chathuranga2016soft}
D.~S. Chathuranga, Z.~Wang, Y.~Noh, T.~Nanayakkara, and S.~Hirai, ``A soft
  three axis force sensor useful for robot grippers,'' in \emph{Intelligent
  Robots and Systems (IROS), 2016 IEEE/RSJ Int. Conf. on}.\hskip 1em plus 0.5em
  minus 0.4em\relax IEEE, 2016.

\bibitem{paulino2017}
T.~Paulino, P.~Ribeiro, M.~Neto, S.~Cardoso, A.~Schmitz, J.~Santos-Victor,
  A.~Bernardino, and L.~Jamone, ``Low-cost 3-axis soft tactile sensors for the
  human-friendly robot vizzy,'' in \emph{Robotics and Automation (ICRA), 2017
  IEEE International Conference on}.\hskip 1em plus 0.5em minus 0.4em\relax
  IEEE, 2017, pp. 966--971.

\bibitem{HEEVER09}
D.~J. van~den Heever, K.~Schreve, and C.~Scheffer, ``Tactile sensing using
  force sensing resistors and a super-resolution algorithm,'' \emph{IEEE
  Sensors Journal}, vol.~9, no.~1, pp. 29--35, Jan 2009.

\bibitem{LEPORA151}
N.~F. Lepora and B.~Ward-Cherrier, ``Superresolution with an optical tactile
  sensor,'' in \emph{Intelligent Robots and Systems (IROS), 2015 IEEE/RSJ
  International Conference on}, Sept 2015, pp. 2686--2691.

\bibitem{LEPORA152}
N.~F. Lepora, U.~Martinez-Hernandez, M.~Evans, L.~Natale, G.~Metta, and T.~J.
  Prescott, ``Tactile superresolution and biomimetic hyperacuity,'' \emph{IEEE
  Transactions on Robotics}, vol.~31, no.~3, pp. 605--618, June 2015.

\bibitem{PIACENZA16_IROS}
P.~Piacenza, Y.~Xiao, S.~Park, I.~Kymissis, and M.~Ciocarlie, ``Contact
  localization through spatially overlapping piezoresistive signals,'' in
  \emph{Intelligent Robots and Systems (IROS), 2016 IEEE/RSJ International
  Conference on}.\hskip 1em plus 0.5em minus 0.4em\relax IEEE, 2016, pp.
  195--201.

\bibitem{PIACENZA17_ICRA}
P.~Piacenza, W.~Dang, E.~Hannigan, J.~Espinal, I.~Hussein, I.~Kymissis, and
  M.~Ciocarlie, ``Accurate contact localization and indentation depth
  prediction with an optics-based tactile sensor,'' in \emph{IEEE Intl. Conf.
  on Robotics and Automation}, 2017.

\bibitem{PONCE14}
R.~D. Ponce~Wong, R.~B. Hellman, and V.~J. Santos, ``Haptic exploration of
  fingertip-sized geometric features using a multimodal tactile sensor,'' in
  \emph{Proc SPIE Defense, Security and Sensing / Sensing Technology and
  Applications “Sensors for Next-Generation Robotics” Conference}, 2014.

\bibitem{WETTELS08}
N.~Wettels, V.~J. Santos, R.~S. Johansson, and G.~E. Loeb, ``Biomimetic tactile
  sensor array,'' \emph{Adv Robot}, vol.~22, no.~8, pp. 829--849, 2008.

\bibitem{ALLEN13}
H.~Dang and P.~K. Allen, ``Grasp adjustment on novel objects using tactile
  experience from similar local geometry,'' in \emph{Intelligent Robots and
  Systems (IROS), 2013 IEEE/RSJ Int. Conf. on}, 2013, pp. 4007--4012.

\bibitem{BEKIROGLU11}
Y.~Bekiroglu, J.~Laaksonen, J.~A. Jorgensen, V.~Kyrki, and D.~Kragic,
  ``Assessing grasp stability based on learning and haptic data,'' \emph{IEEE
  Transactions on Robotics}, vol.~27, no.~3, pp. 616--629, 2011.

\bibitem{SAAL10}
H.~P. Saal, J.~A. Ting, and S.~Vijayakumar, ``Active estimation of object
  dynamics parameters with tactile sensors,'' in \emph{Intelligent Robots and
  Systems (IROS), 2010 IEEE/RSJ International Conference on}, 2010, pp.
  916--921.

\bibitem{TANAKA14}
D.~Tanaka, T.~Matsubara, K.~Ichien, and K.~Sugimoto, ``Object manifold learning
  with action features for active tactile object recognition,'' in
  \emph{Intelligent Robots and Systems (IROS 2014), 2014 IEEE/RSJ International
  Conference on}, 2014, pp. 608--614.

\bibitem{rocsca2010new}
D.~Ro{\c{s}}ca, ``New uniform grids on the sphere,'' \emph{Astronomy \&
  Astrophysics}, vol. 520, p. A63, 2010.

\end{thebibliography}

\end{document}